\documentclass[letterpaper]{article} % DO NOT CHANGE THIS
\usepackage{aaai2026}  % DO NOT CHANGE THIS
\usepackage{times}  % DO NOT CHANGE THIS
\usepackage{helvet}  % DO NOT CHANGE THIS
\usepackage{courier}  % DO NOT CHANGE THIS
\usepackage[hyphens]{url}  % DO NOT CHANGE THIS
\usepackage{graphicx} % DO NOT CHANGE THIS
\usepackage{natbib}  % DO NOT CHANGE THIS AND DO NOT ADD ANY OPTIONS TO IT
\usepackage{caption} % DO NOT CHANGE THIS AND DO NOT ADD ANY OPTIONS TO IT
\usepackage{algorithm}
\usepackage{algorithmic}

\usepackage{newfloat}
\usepackage{listings}
\DeclareCaptionStyle{ruled}{labelfont=normalfont,labelsep=colon,strut=off} % DO NOT CHANGE THIS
\floatstyle{ruled}
\newfloat{listing}{tb}{lst}{}
\floatname{listing}{Listing}
\usepackage{booktabs}       % professional-quality tables
\usepackage{amsfonts}       % blackboard math symbols
\usepackage{nicefrac}       % compact symbols for 1/2, etc.
\usepackage{microtype}      % microtypography
\usepackage{xcolor}         % colors
\usepackage{xspace}
\usepackage{multirow}
\usepackage{subcaption}
\usepackage{amsmath}

\title{RiverScope: High-Resolution River Masking Dataset}
\author{
    Rangel Daroya$^1$ \quad  Taylor Rowley$^1$ \quad  Jonathan Flores$^1$ \quad  Elisa Friedmann$^1$ \quad  \textbf{Fiona Bennitt}$^1$ \\
  \textbf{Heejin An}$^1$ \quad  \textbf{Travis Simmons}$^1$ \quad  \textbf{Marissa Jean Hughes}$^2$ \quad  \textbf{Camryn L Kluetmeier}$^2$  \\
  \textbf{Solomon Kica}$^2$ \quad \textbf{J. Daniel Vélez}$^2$ \quad  \textbf{Sarah E. Esenther}$^3$ \quad  \textbf{Thomas E. Howard}$^3$\\
  \textbf{Yanqi Ye}$^3$ \quad \textbf{Audrey Turcotte}$^4$ \quad  \textbf{Colin Gleason}$^1$ \quad  \textbf{Subhransu Maji}$^1$
}
\affiliations{
$^1$UMass Amherst \quad $^2$UNC Chapel Hill \quad $^3$Brown University \quad $^4$CU Boulder
}

\begin{document}

\maketitle

\begin{abstract}
Surface water dynamics play a critical role in Earth’s climate system, influencing ecosystems, agriculture, disaster resilience, and sustainable development.
Yet monitoring rivers and surface water at fine spatial and temporal scales remains challenging---especially for narrow or sediment-rich rivers that are poorly captured by low-resolution satellite data.
To address this, we introduce RiverScope, a high-resolution dataset developed through collaboration between computer science and hydrology experts.
RiverScope comprises 1,145 high-resolution images (covering 2,577 square kilometers) with expert-labeled river and surface water masks, requiring over 100 hours of manual annotation.
Each image is co-registered with Sentinel-2, SWOT, and the SWOT River Database (SWORD), enabling the evaluation of cost-accuracy trade-offs across sensors---a key consideration for operational water monitoring.
We also establish the first global, high-resolution benchmark for river width estimation, achieving a median error of 7.2 meters---significantly outperforming existing satellite-derived methods.
We extensively evaluate deep networks across multiple architectures (e.g., CNNs and transformers), pretraining strategies (e.g., supervised and self-supervised), and training datasets (e.g., ImageNet and satellite imagery). 
Our best-performing models combine the benefits of transfer learning with the use of all the multispectral PlanetScope channels via learned adaptors.
RiverScope provides a valuable resource for fine-scale and multi-sensor hydrological modeling, supporting climate adaptation and sustainable water management.
\end{abstract}

\begin{links}
    \link{Code}{https://github.com/cvl-umass/riverscope-models}
    \link{Dataset}{https://github.com/cvl-umass/riverscope}
    % \link{Extended version}{https://arxiv.org/abs/2509.02451}
\end{links}

\begin{figure*}[!t]
    \centering
    \includegraphics[width=0.9\linewidth]{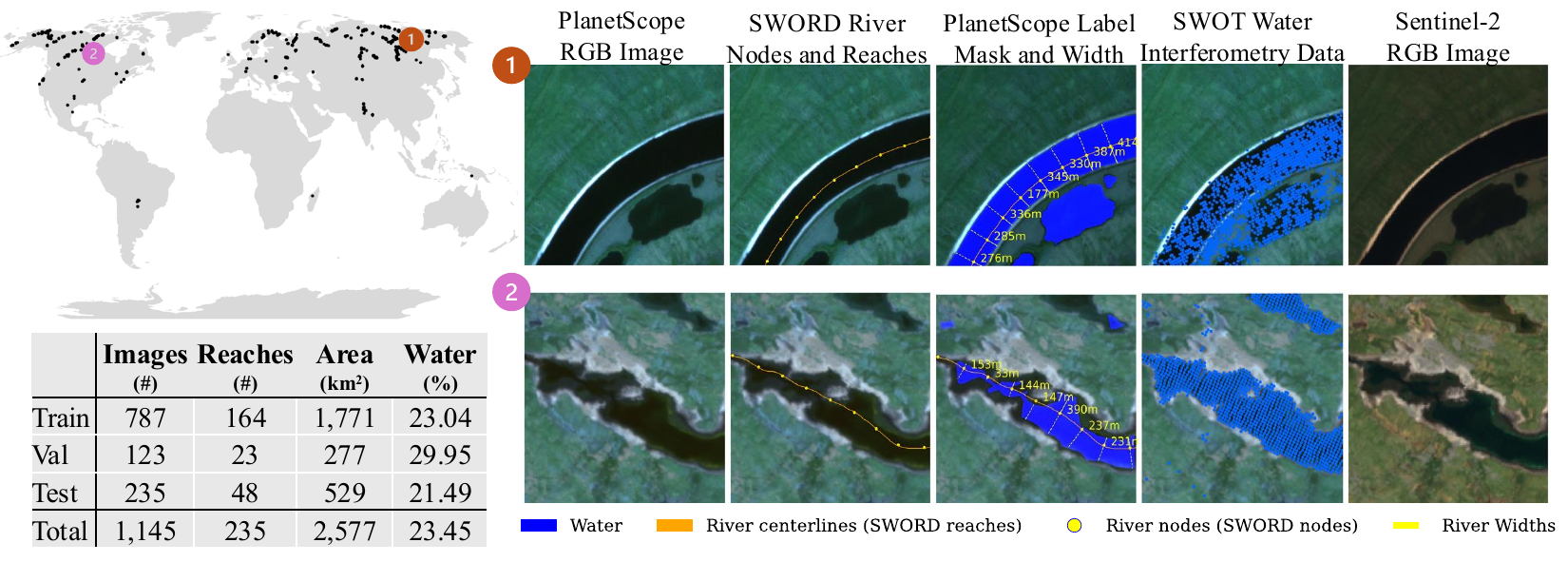}
    \caption{RiverScope presents a global, high-resolution satellite image dataset focused on rivers using PlanetScope~\cite{planetlabs} and co-registered with SWOT~\cite{vinogradova2025new}, SWORD~\cite{altenau2021surface}, and Sentinel-2~\cite{esa2022sentinel}. 
    To the left we show the distribution and splits of our expert-labeled dataset, covering various geographic and hydrological contexts.
    }
    \label{fig:overview}
\end{figure*}

\section{Introduction}
\label{sec:intro}
Global surface waters form the circulatory system of the climate, playing a vital role in transporting water and the materials it carries across the planet~\cite{alsdorf2007measuring}. These dynamics are shaped not only by natural phenomena like floods and erosion but also by human activities like dam construction and irrigation~\cite{yang2022impact}. Understanding and mapping surface water dynamics is increasingly important for climate resilience and environmental sustainability due to their role in agriculture~\cite{tian2015modeling}, hydropower~\cite{wasti2022climate}, ecological services~\cite{zhao2003ecosystem}, urban planning~\cite{ellis2013sustainable}, and transportation~\cite{opher2010factors}.

While the hydrology community has made progress in locating and mapping surface water (i.e., hydrography)~\cite{valman2024ai}, characterizing water dynamics like extent and flow remains challenging. Satellite imagery is effective at capturing water extent, but properties such as flow are not directly observable~\cite{feng2019comparing}. Stream gauges, which directly measure flow, are sparse and insufficient to provide a global view~\cite{gleason2017crossing}.

These limitations have driven the launch of the Surface Water and Ocean Topography (SWOT) mission~\cite{biancamaria2016remote}, which uses radar interferometry to measure surface water elevation with unprecedented precision~\cite{vinogradova2025new}. However, river discharge (flow) is still inferred, a calculation that critically depends on accurate river width measurements~\cite{bjerklie2018satellite}. To support this, datasets like the SWOT River Database (SWORD)~\cite{altenau2021surface} were developed, but they rely on width estimates from coarser Landsat imagery~\cite{allen2018global}. 

Spatial resolution remains a key bottleneck, especially for small rivers and fine-scale changes~\cite{filippucci2022sentinel}. Sentinel-2~\cite{esa2022sentinel} and Landsat~\cite{eros2020_landsat_l2_c2} imagery (10-30m/pixel) often miss these features~\cite{flores2024mapping}, limiting the accuracy of river models. This motivates the need for higher-resolution, expertly labeled datasets.

We introduce RiverScope, a densely annotated, global-scale dataset of high-resolution (3~m/pixel) PlanetScope~\cite{planetlabs} imagery of rivers and adjacent water bodies. The dataset covers 2,577 $km^2$, comprising 500x500-pixel images sampled across diverse geographic and hydrological contexts. Each image is co-located with Sentinel-2, SWOT, and SWORD data within a $\pm$12-hour window (see Figure~\ref{fig:overview}). This unique alignment enables the comparison of sensor performance on hydrologically relevant tasks, quantifying trade-offs between accuracy and cost---a critical consideration for scaling monitoring systems at agencies like NASA or ESA. Surface water masks were created by hydrology and machine learning experts through 100 hours of manual annotation, ensuring high-quality ground truth data.  We leverage this dataset to establish benchmarks using hydrologically relevant evaluation metrics.

Rivers are an important, underexplored domain for machine learning, with complex natural features distinct from structured objects in typical urban datasets. RiverScope captures complex hydrological features such as riverbanks, sandbars, and varying morphologies---elements that are often lost in coarser datasets. By leveraging high-resolution multispectral imagery, models can better delineate river boundaries, improving width estimates essential for discharge modeling.

\begin{figure}[!t]
    \centering
    \includegraphics[width=0.95\linewidth]{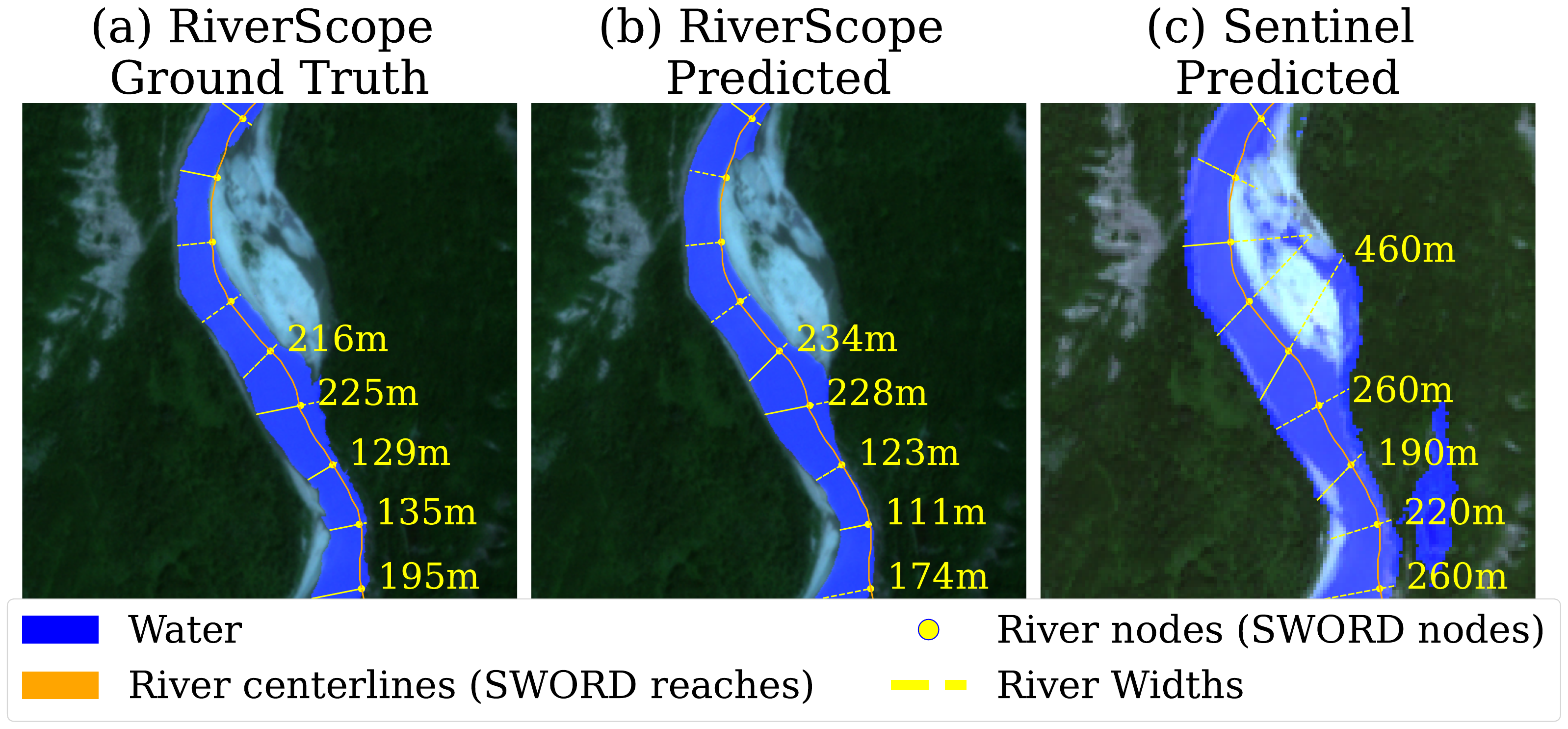}
    \caption{RiverScope can be used to precisely segment rivers and water bodies (a-b). Existing low-resolution images like Sentinel (c) tend to over segment narrow rivers, inflating river width estimates due to less detail in the images.
    Yellow dots mark SWORD nodes; the orange line represents a section of the SWORD reach used as the river centerline.
    }
    \label{fig:planet-vs-sentinel-segmentation}
\end{figure}

Our benchmark serves two key purposes:
\begin{enumerate}
\item \textbf{Water segmentation.} We evaluate diverse deep learning models on PlanetScope imagery, including ImageNet-pretrained models and remote sensing-specific methods. We show that lightweight input adapters using all the multispectral PlanetScope bands outperform RGB-only models. Our models outperform established baselines including NDWI~\cite{mcfeeters1996use} and custom CNN architectures~\cite{valman2024ai}. PlanetScope's fine spatial resolution enables detection of detailed river structures often missed by lower-resolution sensors like Sentinel (see Figure~\ref{fig:planet-vs-sentinel-segmentation}). Models trained on coarse imagery often struggle with precise boundary delineation, highlighting the potential of fusing complementary modalities in our dataset to improve performance of Sentinel estimates.

\item \textbf{River width estimation.} We introduce a benchmark for river width---a hydrologically meaningful metric poorly captured by pixel-level segmentation metrics like IoU. Our best model, trained on PlanetScope imagery, achieves a median width error of 7.2 meters, substantially outperforming the prior state-of-the-art of 30 meters~\cite{valman2024ai}. Landsat- and Sentinel-derived widths show median errors of 45.0 and 39.0 meters, respectively. Notably, we also provide the first evaluation of SWOT-derived widths, which exhibit a median error of 41.9 meters at ``nodes" spaced every 200 meters along the river.
\end{enumerate}

By publicly releasing RiverScope, we aim to foster machine learning research on domain-specific challenges in river monitoring---problems with direct implications for climate adaptation and sustainable water management.

\section{Related Work}
\label{sec:related-work}

\paragraph{Satellite image datasets.} 
Previous works have introduced large-scale datasets based on satellites with global coverage, such as Sentinel and Landsat~\cite{bastani2023satlaspretrain, claverie2018harmonized}. For example, SatlasPretrain~\cite{bastani2023satlaspretrain} presents a diverse dataset of Sentinel and NAIP~\cite{usda_naip_imagery} imagery labeled for object detection, scene classification, and segmentation. EuroSAT~\cite{helber2019eurosat} and BigEarthNet~\cite{sumbul2019bigearthnet} provide Sentinel-based datasets for image classification, while NASA periodically releases segmentation labels through the Harmonized Landsat-Sentinel program~\cite{jones2019improved}, including water body mapping products~\cite{daroya2025improving}.

While Sentinel and Landsat offer broad spatial and temporal coverage, their resolutions (10~m for Sentinel, 30~m for Landsat) limit their utility for fine-grained hydrological analysis, such as tracking narrow river dynamics. High-resolution datasets remain limited and are often focused on urban or industrial settings, relying on aerial imagery with sparse global coverage and infrequent revisit rates~\cite{christie2018functional_fmow}.

The PlanetScope mission~\cite{planetlabs} addresses these limitations by offering globally available, high-resolution (3~m per pixel), near-daily imagery. This has enabled recent work on surface water monitoring using PlanetScope data~\cite{valman2024ai, flores2024mapping}. However, these efforts have largely focused on specific regions and \emph{do not} publicly release annotated datasets.

To address this gap, we introduce RiverScope, a densely labeled, global-scale dataset of high-resolution PlanetScope imagery for water segmentation focused on rivers. While \citet{valman2024ai} presents a similar large-scale water segmentation dataset, RiverScope extends this line of work by incorporating co-registered measurements from complementary sensors—including Sentinel~\cite{esa2022sentinel}, Surface Water and Ocean Topography (SWOT)~\cite{biancamaria2016remote}, and SWOT River Database (SWORD)~\cite{altenau2021surface}—and by \emph{openly} releasing the annotated data.

The recently launched SWOT satellite provides interferometric measurements of surface water elevation and extent, while SWORD offers reach-level hydrologic variables such as river width and slope (See~\S~\ref{sec:data-construction} for details). By aligning these with PlanetScope and Sentinel imagery, RiverScope enables benchmarking across multiple tasks and facilitates evaluating cost-accuracy trade-offs of different sensors---a key consideration for scaling monitoring systems.

\paragraph{Satellite image models.} 
While early studies in satellite image modeling focused primarily on low-resolution data~\cite{bastani2023satlaspretrain, manas2021seasonal_seco, daroya2025improving}, a growing body of work now targets applications requiring high-resolution imagery~\cite{flores2024mapping}. For instance, \citet{flores2024mapping} used PlanetScope imagery to segment headwater streams---small, often unnamed tributaries that require fine spatial details. Beyond stream segmentation~\cite{valman2024ai}, PlanetScope has been applied to river discharge estimation, reservoir monitoring, and water quality assessment~\cite{wang2022mapping, flores2024mapping}.

Building on prior work~\cite{valman2024ai}, we focus on water segmentation using PlanetScope, and further extend to estimating river widths and validating outputs of complementary sources such as Sentinel, SWOT, and SWORD. Despite increased interest in high-resolution applications, there has been limited exploration of pretraining strategies (e.g., ImageNet~\cite{deng2009imagenet}, CLIP~\cite{radford2021learning_clip}, MoCov3~\cite{chen2021empirical_mocov3}) and segmentation architectures~\cite{ronneberger2015u, ranftl2021vision} specifically tailored for this domain. Many prior approaches rely on randomly initialized convolutional models~\cite{valman2024ai} or simple thresholding methods~\cite{mcfeeters1996use}.

In this work, we benchmark a range of segmentation architectures and pretraining methods for high-resolution river segmentation and width estimation. Across all models and training strategies, higher-resolution imagery delivers higher accuracy, highlighting the clear advantage of fine spatial detail for both river segmentation and width estimation.

\section{RiverScope Construction}
\label{sec:data-construction}

\subsection{Data Sources}
We collect satellite image data from the PlanetScope and the Sentinel-2 constellation of satellites.
PlanetScope data were selected to cover areas that spatially and temporally intersect with SWOT, SWORD, and Sentinel-2. 
This is done to give access to multiple observation modalities in the same location and time, while enabling cross-validation across different sources.
PlanetScope images were used for annotating water masks to capture fine details that high-resolution imagery better reveals.
We discuss the collection process for each of the data sources below (more details in \S~\ref{subsec:supp-riverscope-details} (Appendix)).

\paragraph{Surface Water and Ocean Topography (SWOT)} is a satellite mission launched by NASA and CNES in December 2022, and orbits with a Ka-band Radar Interferometer (KaRIn) that revisits almost the entire globe every 21 days.
SWOT produces a three-dimensional point cloud of the Earth's water surfaces by timing microwave pulses and comparing the phase returned to its twin antennae.
Ground processing first classifies each radar sample as water or non-water, then aggregates these classifications along rivers at fixed nodes (i.e., points every 200~m) and reaches (i.e., line segments around 10~km long).
Hydrological variables are derived from these aggregates such as effective width (water area divided by effective length), surface elevation, and slope.

Since Ka-band radar is less sensitive to clouds, SWOT provides large-scale, repeatable monitoring of rivers, lakes, and reservoirs wider than about 50~m worldwide, offering an alternative to traditional gauge-based systems.
Previous studies~\cite{yao2025swot} show SWOT can adequately resolve rivers, but mask quality is affected by factors such as vegetation, wetlands, and nearby urban areas. 
Our dataset provides the first systematic evaluation of river width estimates derived from SWOT on a global scale. 

\paragraph{SWOT River Database (SWORD)} is a global database of rivers designed to support SWOT river products. 
It integrates multiple data sources including river widths estimated from Landsat imagery (30~m/pixel resolution), river flow characteristics extracted from digital elevation maps~\cite{yamazaki2019merit}, and identified human-made obstructions along river paths.
SWORD structures rivers using points (\textbf{nodes}) spaced approximately 200~m apart, connected by line segments (\textbf{reaches}) around 10~km long (see Figure~\ref{fig:planet-vs-sentinel-segmentation}).
These nodes and reaches serve as reference points from which SWOT data are obtained.
SWORD also provides river width estimates at the node level, which are part of our dataset.

\paragraph{Sentinel-2} data from the European Space Agency (ESA) provides 10~m/pixel to 60~m/pixel multispectral imagery, depending on the spectral band (the red, blue, green, and near infrared (NIR) bands have 10~m/pixel resolution).
We download tiles that correspond to the same location and approximate time as PlanetScope imagery, while selecting tiles with the least cloud cover.
Although water segmentation labels are available from WorldCover~\cite{van2021esa_worldcover}, these only provide data for 2020 and 2021 with a 10~m/pixel resolution.
There are 13 spectral bands available for Sentinel imagery.
We include all available spectral bands as part of our dataset.

\paragraph{PlanetScope} has a constellation of satellites collecting 3~m/pixel resolution image data with a revisit frequency of 1 day.
Each image has 4 bands corresponding to the red, green, blue, and NIR bands.
Images of size 500$\times$500 pixels were collected from 2023 to 2024 to be within the same time frame ($\pm$ 12 hours) as SWOT. 
River sites spanning a broad range of widths were selected using width estimates from other sensors (e.g., SWORD) and were limited to SWOT's fast sampling orbit (i.e., the path where SWOT makes frequent measurements) to maximize the number of spatially intersecting samples.
Figure~\ref{fig:overview} shows the distribution of the obtained 1,145 images from different geographic locations.
Images with no cloud cover were selected to ensure optimal visibility of rivers.
We manually label water pixels in all these images to cover an area of 2,577 $km^2$ (see \S~\ref{sec:data-labeling}). 
We publicly release normalized versions (i.e., min-max normalized pixel values per image) of the multispectral data. Links to the original PlanetScope products with raw values are also available, and can be purchased directly from~\cite{planetlabs}.
Models presented here are trained on the normalized data.

\subsection{PlanetScope Data Labeling}
\label{sec:data-labeling}
Unlike Sentinel images, PlanetScope images provide higher temporal and spatial resolution, enabling more detailed fine-grained image analyses.
Consequently, these higher-resolution images were selected for manual annotation.
Each PlanetScope image was manually annotated by one of 15 hydrology and river experts using a multi-scale pixel-wise annotator~\cite{tangseng2017looking}.
Annotators were provided with the RGB image of the river to label, corresponding SWORD reaches and nodes overlaid on the images, a zoomed-out contextual view surrounding the target area, and a Google Maps link of the location.
These supplementary information were included to enhance context and accuracy in labeling.
Annotators also had interactive abilities such as zooming in and out of the target area, and the option of using a variety of labeling tools.
Labels can be provided using a paintbrush, a polygon selector, and a superpixel selection tool that clusters similar pixels together.
These interactive features streamlined the annotation process, ensuring comprehensive and efficient label coverage.
A detailed description and visualization of the tool is provided in \S~\ref{subsec:supp-data-labeling} (Appendix).

Each annotator provided labels for all water pixels in the image, differentiating between river and non-river water. 
However, since the primary objective was detailed river analysis, most collected satellite imagery predominantly featured river water.
Thus, our focus was on accurate segmentation of all water pixels to facilitate further river-specific analyses.
Annotating each image required at least 5 minutes, with more complicated river configurations taking more time, totaling more than 100 hours of expert annotation effort for all collected PlanetScope images.

\subsection{Data splits}
We split the data geographically so the training data cover different areas than the validation and test.
River reaches (i.e., 10~km river sections) were separated so that each river appears exclusively in one data split.
Specifically, the dataset contains 164 unique river reaches in the training set, 23 in the validation set, and 48 in the test set.
The splits were constructed to have a similar average number of water pixels per image, ensuring balanced representation (see Figure~\ref{fig:overview}).

\subsection{Tasks}
\label{sec:tasks}
Using the available data in RiverScope, we evaluate performance of existing architectures and pretraining methods on water and river-specific tasks. In particular, we investigate (1) water segmentation performance and (2) river width estimation performance of existing models.

\paragraph{Water segmentation.} We investigate water segmentation performance on high-resolution satellite imagery using the labeled PlanetScope images. 
RiverScope models are trained on the PlanetScope train set, with hyperparameter selection done using F1 score on the corresponding validation set. 
All models are then evaluated on the held out PlanetScope test set.
To evaluate Sentinel models, we predict on the corresponding Sentinel images taken in the same location and time as the PlanetScope test images.
The F1 score metric is used to evaluate model accuracy, since it effectively accounts for class imbalance caused by the relatively small proportion of water pixels (around 20\%, see Figure~\ref{fig:overview}).

\paragraph{River width estimation.} For each node defined in SWORD, the ground truth river width is derived from PlanetScope water segmentation labels (see Figure~\ref{fig:planet-vs-sentinel-segmentation}).
Given a satellite image, the corresponding river reach that spatially intersects the image is retrieved from SWORD.
For each node along the river reach, the local slope is estimated, and a perpendicular (orthogonal) line is computed.
River widths are then calculated by counting the number of water pixels along this orthogonal line and multiplying by the image resolution (3~m/pixel for PlanetScope)~\cite{yang2019rivwidthcloud}.
These measurements serve as the ground truth river widths for SWORD nodes and provide the reference for evaluating other methods.

To evaluate performance, we use bias, \%~bias, mean absolute error, and median absolute error.
The bias is computed by subtracting the ground truth widths $y_i$ from the predicted widths $\hat{y}_i$ and getting the average ($\frac{1}{N} \sum_{i=1}^N \left( \hat{y}_i-y_i \right)$).
The \% bias is computed by dividing the ground truth width for each prediction: $\frac{1}{N} \sum_{i=1}^N \frac{\hat{y}_i - y_i}{y_i}$.
The mean and median absolute errors are computed as the mean and median of $|y_i - \hat{y}_i| \ \forall i$, respectively.
Model predictions are compared against width estimates from Landsat (via SWORD), Sentinel, and SWOT across all river nodes. 
The evaluation is limited to rivers with widths of 500~m or less since the labeled images are 500x500 pixels in size and do not reliably capture wider rivers.
A total of 445 nodes were used exclusively for evaluation.

\section{RiverScope Models}
\label{sec:models}

\subsection{Training Details}
\label{subsec:training-details}
We evaluate 27 models on RiverScope tasks based on varying segmentation models, backbones, and pretraining methods~\cite{Iakubovskii:2019} to see the effect of training with our dataset across different settings.
We experiment on four semantic segmentation models covering different ways of handling features and multi-scale information: FPN~\cite{long2015fully}, DeepLabv3~\cite{chen2017deeplab}, UNet~\cite{ronneberger2015u}, DPT~\cite{ranftl2021vision}.
For each segmentation model, we then experiment with different backbones and available pretrained weights which are the basis of extracting features for each segmentation model.
FPN, DeepLabv3, and UNet are applied to CNN and Swin-based backbones, while DPT is applied to ViT-based backbones.
ResNet50 (RN50)~\cite{heDeepResidualLearning2016}, MobileNetv2 (MV2)~\cite{sandler2018mobilenetv2}, Swin-T~\cite{liu2021swin}, Swin-B~\cite{liu2021swin}, ViT-B/16~\cite{dosovitskiy2020vit}, and ViT-L/16~\cite{dosovitskiy2020vit} backbones are used.
Pretraining methods also vary from supervised methods using SatlasNet~\cite{bastani2023satlaspretrain} and ImagetNet1k~\cite{deng2009imagenet}, and self-supervised methods using SeCo~\cite{manas2021seasonal_seco}, MoCov3~\cite{chen2021empirical_mocov3}, CLIP~\cite{radford2021learning_clip}, Prithvi~\cite{jakubik2310foundation_prithvi}, and DINO~\cite{caron2021emerging}.
Among these, SeCo, SatlasNet, and Prithvi use satellite images for pretraining.
Each of these configurations serves as the starting point before further fine-tuning with RiverScope.

All models are trained on the PlanetScope labeled images from RiverScope. Binary cross entropy loss (Eq.~\ref{eq:bce}) is used to train all models. A predicted segmentation mask $\hat{\mathbf{Y}}_i \in \mathbb{R}^{W\times H}$ with pixels $\hat{y}_{i,(j,k)}$ is compared with the ground truth mask $\mathbf{Y}_i \in \mathbb{R}^{W\times H}$ using Eq.~\ref{eq:bce}. The Adam optimizer is used for training with the learning rate chosen from $10^{-1}$ to $10^{-6}$ based on the validation set performance.

\begingroup
\fontsize{8pt}{2pt}\selectfont
\begin{multline}
    \mathcal{L}_{bce}\left(\mathbf{Y}_i, \hat{\mathbf{Y}}_i \right) = \frac{1}{WH} \sum_{j=1}^W \sum_{k=1}^H - \Biggl(y_{i,(j,k)} \log \hat{y}_{i,(j,k)} \\+ \left(1-y_{i,(j,k)} \right) \log\left(1-\hat{y}_{i,(j,k)} \right) \Biggr)
    \label{eq:bce}
\end{multline}
\endgroup

\begin{figure}[t]
    \centering
    \includegraphics[width=\linewidth]{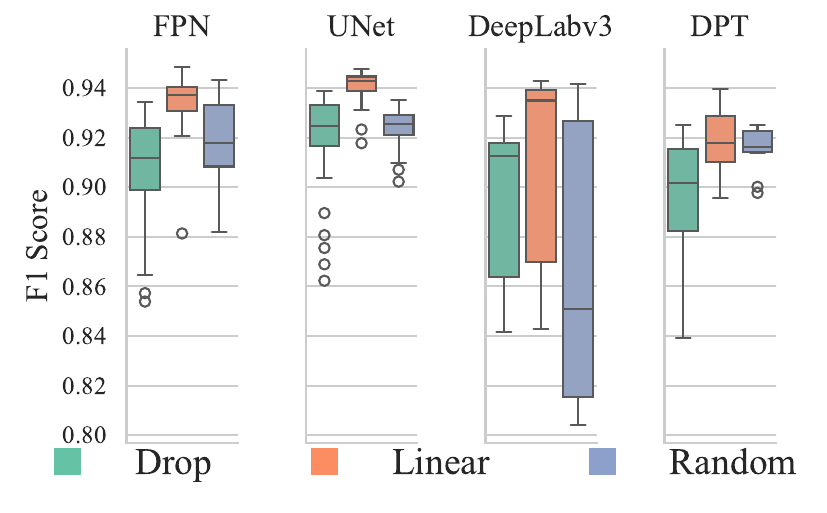}
    \caption{Segmentation performance of different methods for adapting a 4-channel satellite image to RGB to utilize existing RGB pretrained models. `Drop' refers to dropping the NIR channel, `Linear' refers to applying a linear layer to convert 4 channels to 3 channels, and `Random' refers to training 4-channel models without any pretraining applied. We find that linearly projecting the input from 4-channels to 3-channels worked best (raw numbers in Table~\ref{table:supp-adaptor-quantitative} (Appendix)).}
    \label{fig:adapter-results}
\end{figure}

\subsection{Baseline Methods}
To evaluate (1) the effect of training on high-resolution satellite images and (2) the advantage of using pretrained models on the tasks, we compare results on several existing baselines.
For (1), we compare performance against the same models trained on Sentinel data which have lower resolution.
For (2), we compare on two models: a recent river width estimation model trained on PlanetScope data by \citet{valman2024ai}, and a widely used water segmentation algorithm NDWI~\cite{mcfeeters1996use}.

\noindent\textbf{Sentinel models} are trained on sampled Sentinel data from 2020-2021 with WorldCover~\cite{van2021esa_worldcover} labels for water segmentation. The same segmentation models, backbones, and pretraining methods from \S~\ref{subsec:training-details} are also used to evaluate the effect of using low-resolution data on the downstream tasks. 
Multispectral data are used as input for all Sentinel models.
We include more details in the Appendix.

\noindent\textbf{PlanetScope CNN}~\cite{valman2024ai} is a recent work that similarly labeled rivers from PlanetScope images. They use a randomly initialized custom CNN that was trained to segment water from satellite images, and was designed to take four bands as input.

\noindent\textbf{PlanetScope NDWI}~\cite{mcfeeters1996use} uses the Normalized Difference Water Index (NDWI) to detect water from satellite images, which is widely used in remote sensing. NDWI is a value between -1 and 1 that uses the difference in the green and NIR reflectance values to detect water (Eq.~\ref{eq:ndwi}), since water reflects almost no NIR. To find the threshold $t$ for water (i.e., $\text{NDWI} > t$), Otsu thresholding~\cite{otsu1975threshold} is applied.

\begin{equation}
    \text{NDWI} = \frac{\text{green} - \text{NIR}}{\text{green} + \text{NIR}}
    \label{eq:ndwi}
\end{equation}

\section{Experiments}
\label{sec:experiments}

\begin{figure}[t]
    \centering
    \includegraphics[width=\linewidth]{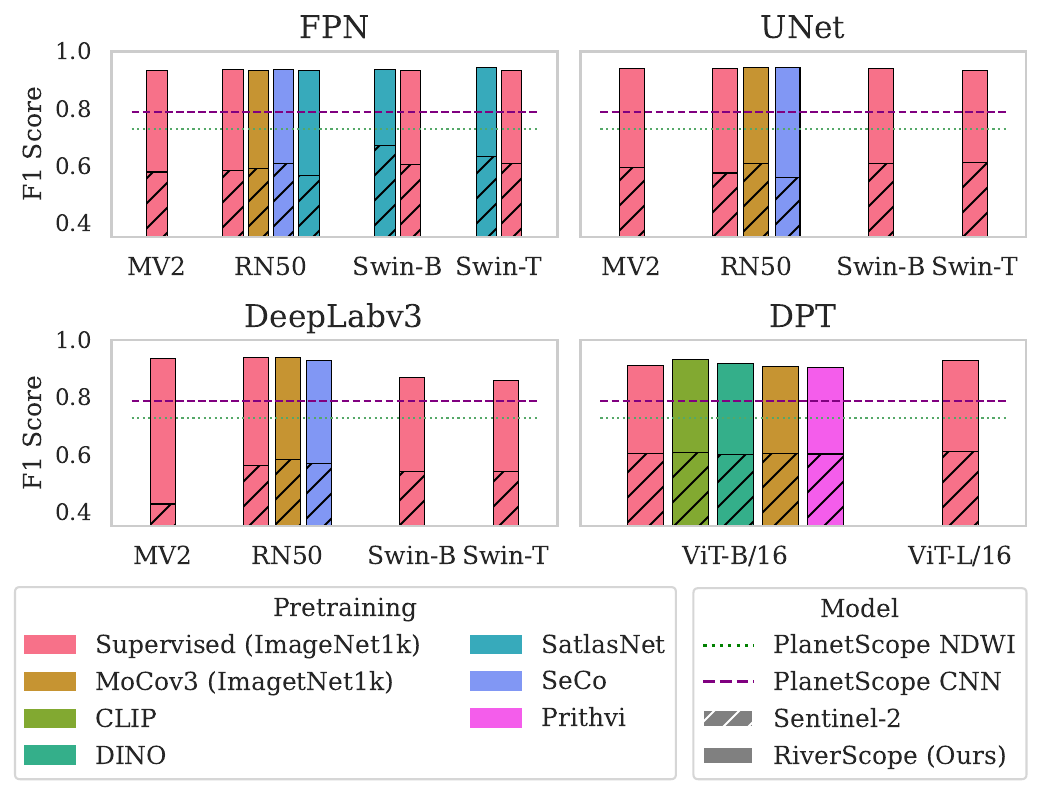}
    \caption{RiverScope trained models more accurately segment river pixels compared to Sentinel trained models. Each subplot shows the average F1 score improvement of a given segmentation model across multiple runs. For each architecture and pretraining combination, hatched bars represent the performance of Sentinel trained models, while solid bars represent the performance of RiverScope trained models. 
    We show raw numbers in Table~\ref{table:river-segmentation-results},~\ref{table:supp-river-segmentation-results-planet-baselines} (Appendix). 
    }
    \label{fig:planet-segmentation-results}
\end{figure}

\subsection{Water Segmentation}
\label{subsec:expt-water-segmentation}
\noindent\textbf{Linear adaptation of multispectral data to RGB pretrained models yields optimal performance.} 
Since RiverScope labeled image data is composed of four channels, we evaluate the most effective way to use pretrained models that take three-channel RGB images as input.
We look into (1) dropping the additional NIR channel to effectively keep only RGB (\textbf{Drop}), (2) linearly projecting the 4-channel input to 3-channel (\textbf{Linear}), and (3) training 4-channel models from scratch without using pretrained models (\textbf{Random}).
Figure~\ref{fig:adapter-results} shows the performance across these different settings.
Our results indicate that the linear adaptor yields the best performance in terms of average F1 score across different architectures, outperforming other methods. 
There is also generally less variance in the performance of `Linear'.
This suggests that leveraging RGB pretrained models leads to better performance than training from scratch, even when using additional channels.
Based on these findings, we adopt the linear adaptor for all subsequent experiments.

\noindent\textbf{Water segmentation improves by training on high-resolution data.} 
Figure~\ref{fig:planet-segmentation-results} displays the difference in performance when models are trained on high-resolution RiverScope data compared to models trained on low-resolution Sentinel data. 
RiverScope's best model is a SeCo pretrained RN50 UNet, whereas Sentinel's is a Satlas pretrained Swin-B FPN.
Performance is likely positively affected by the presence of more details in high-resolution images (Figure~\ref{fig:planet-vs-sentinel-segmentation}) that enable models to better distinguish between water and non-water.
For example, sand bars around rivers could be better seen if more details were present in the image.
In addition, since high-resolution images can more precisely define the boundaries of water bodies, models trained on high-resolution data can learn these boundaries with better accuracy. 
This is supported by additional results in Table~\ref{table:river-segmentation-results} (Appendix) showing the very low precision of Sentinel models (see Figure~\ref{fig:supp-visualization-results} (Appendix) for visualizations).

\noindent\textbf{Using pretrained models improve segmentation performance.} 
While PlanetScope CNN and PlanetScope NDWI also use high-resolution multispectral images, their performance is still lower compared to any of the RiverScope models that utilize transfer learning (Figure~\ref{fig:planet-segmentation-results}). 
Due to the rule-based nature of NDWI, it is sensitive to man-made land features and generally overestimates water boundaries which leads to falsely identifying pixels as water~\cite{xu2006modification} (see Table~\ref{table:supp-river-segmentation-results-planet-baselines}, Figure~\ref{fig:supp-error-visualization} (Appendix)).
PlanetScope CNN tends to miss a significant number of water pixels, resulting in more false negatives and lower recall (see Figure~\ref{fig:supp-error-visualization}, Figure~\ref{fig:supp-visualization-results} (Appendix)).
Pretraining on large datasets such as ImageNet and SatlasPretrain can speed up the process of learning image features and result in better performance.

\noindent\textbf{RiverScope and Sentinel trained models can be improved by reducing false positives.} 
Table~\ref{table:river-segmentation-results} (Appendix) shows that for both types of trained models, precision trails behind recall, confirming that false positives---not missed detections---are the performance bottlenecks.
Although a Sentinel model generally performs poorly compared to a RiverScope model, the former's recall is competitive, indicating that it can locate water pixels properly; it mainly struggles with precision.
As detailed in Table~\ref{table:supp-false-positive-seg} (Appendix), most false positives come from `Tree Cover', `Grassland', and `Herbaceous Wetland' which typically occur along the boundary of rivers, and can look similar to water.
Augmenting the dataset with additional examples of these three land cover types could potentially improve both Sentinel and RiverScope trained models.
Alternatively, aggregating predictions across multiple dates for the same location might resolve false positives, since it is likely that a falsely identified water-like pixel would only appear as water in one time snapshot.

\subsection{River Width Estimation}
\begin{table}[!t]
    \setlength{\tabcolsep}{2.5pt}
    \small
    \centering
   \begin{tabular}{ c | l |r r r r }
        \toprule	
        PS$^{\dagger}$& & Mean & Median & Bias & \% Bias \\
        \midrule
        &SWOT & 94.4 & 41.9 & 28.2 & 38.0 \\
        &GRWL (SWORD/Landsat) & 77.0 & 45.0 & -6.9 & 25.0 \\
        &Sentinel & 152.8 & 39.0 & 119.8 & 202.1 \\
        \midrule
        \checkmark &NDWI~\cite{mcfeeters1996use} & 194.6 & 51.0 & 160.2 & 176.3 \\    
        \checkmark &CNN~\cite{valman2024ai} & 87.2 & 30.0 & 39.3 & 14.0 \\
        \checkmark &RiverScope (Ours) & \textbf{15.3} & \textbf{7.2} & \textbf{5.7} & \textbf{11.4} \\
         \bottomrule
     \end{tabular}
     \caption{River width estimation errors (m) for the best performing Sentinel (Satlas pretrained Swin-B FPN) and RiverScope (ImageNet pretrained RN50 FPN) models compared to other baselines. The RiverScope trained model has the lowest errors overall. Raw numbers are in Table~\ref{table:addtl-river-width-estimation-results} (Appendix).
    $^{\dagger}$PS:~PlanetScope is used as input}
    \label{table:river-width-estimation-results}
\end{table}

\begin{figure}[t]
    \centering
    \includegraphics[width=\linewidth]{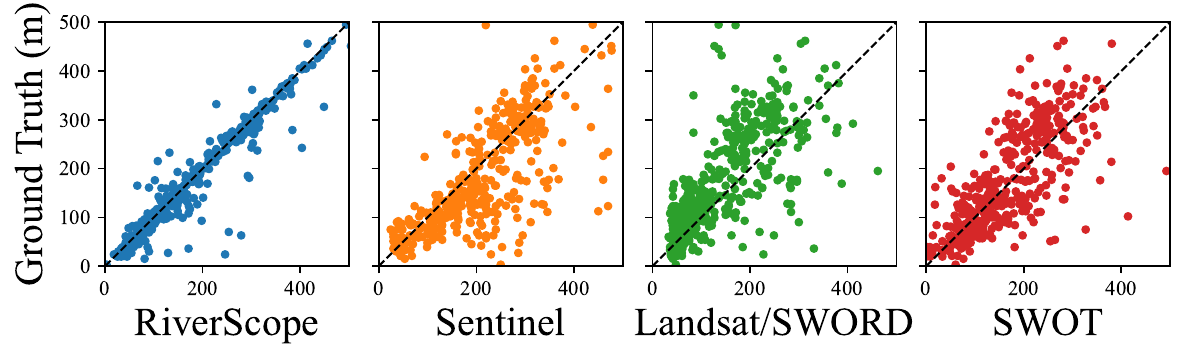}
    \caption{Distribution of width estimates. The RiverScope model predicted widths that cluster closely to the $y=x$ line.
    }
    \label{fig:scatter-river-width-estimation-results}
\end{figure}

\begin{figure}[t]
    \centering
    \includegraphics[width=\linewidth]{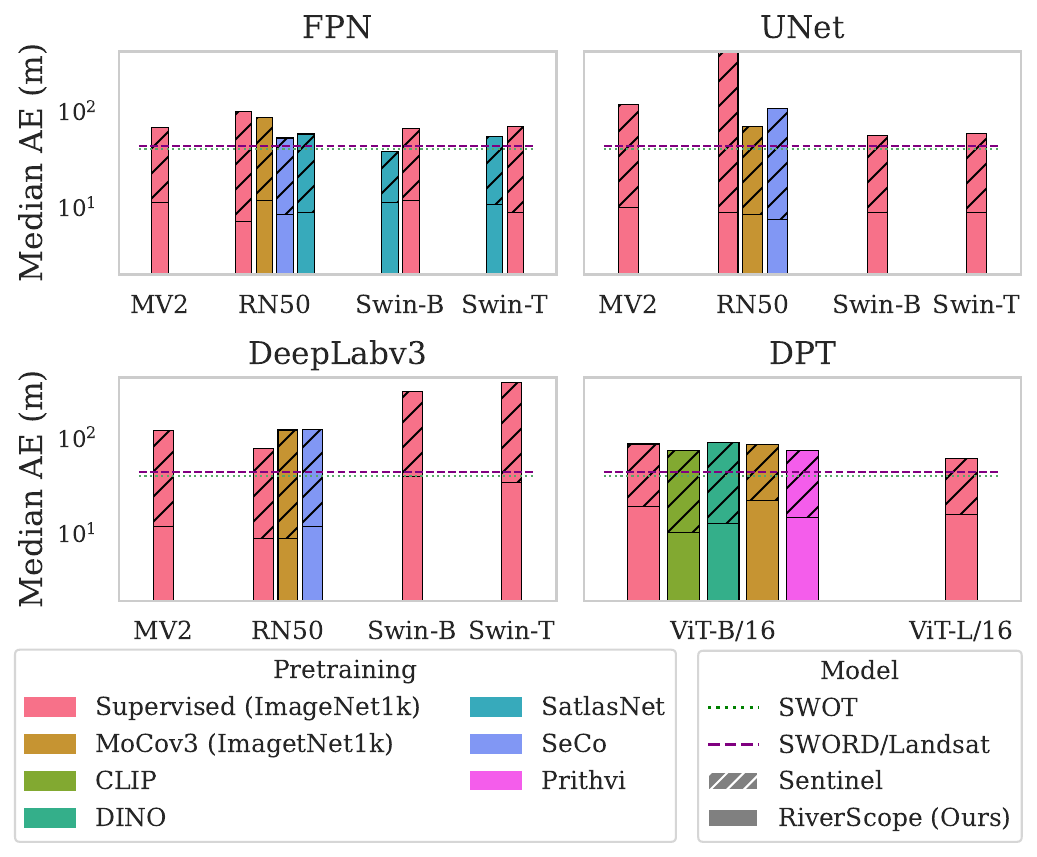}
    \caption{RiverScope trained models improve performance on river width estimation. Each subplot shows the median absolute error (m) for a given segmentation model. For each architecture and pretraining combination, the hatched bar is the performance of a Sentinel trained model, while the solid bar is of a RiverScope trained model. We additionally show estimates from sensors SWOT and Landsat. Across the board, we see significantly lower error when using a model trained with high-resolution RiverScope images. We show the raw numbers in Table~\ref{table:addtl-river-width-estimation-results} (Appendix).
    }
    \label{fig:river-width-results}
\end{figure}

\noindent\textbf{RiverScope trained models have lower river width estimation error than their low-resolution counterparts.} 
Table~\ref{table:river-width-estimation-results} shows the RiverScope trained model has the lowest error, while Figure~\ref{fig:scatter-river-width-estimation-results} shows it overpredicts less than other models.
This improvement can be linked to the better water segmentation performance of the RiverScope model (Figure~\ref{fig:planet-segmentation-results}).
With more accurate identification of river waters (as a result of the finer details and the more precise water boundaries in high-resolution satellite images), the estimated widths are also likely to be more precise.
Figure~\ref{fig:river-width-results} shows comparisons across different architectures and settings, showing that in all configurations, training with RiverScope leads to superior river width estimation performance.

\noindent\textbf{Higher-resolution training yields more accurate predictions.}
The Riverscope model trained on 3~m PlanetScope imagery (highest resolution among models in Table~\ref{table:river-width-estimation-results}), achieves the best width estimation performance.
River width is estimated by counting water pixels along the orthogonal (Figure~\ref{fig:planet-vs-sentinel-segmentation}), so a one pixel misclassification at each riverbank side gets an error of $2\sqrt{2} \Delta$ if the orthogonal cuts the pixel diagonally ($2\Delta$ if edge-aligned), where $\Delta$ is a sensor's resolution.
For Sentinel this is 28~m (20~m if edge-aligned), so the observed 39~m median absolute error (Table~\ref{table:river-width-estimation-results}) implies roughly 2 pixels of error.
PlanetScope's 3~m/px resolution lowers this to 8.5~m (6~m), and our trained model achieves 7.2~m error---essentially a 1 pixel error.
RiverScope models are not only more precise due to the increased spatial resolution, but also better at resolving misclassifications of water.

\subsection{Future Directions}
\noindent\textbf{Sentinel predictions could potentially be further improved by removing outliers.}
While Sentinel has higher resolution than Landsat (10~m/px vs 30~m/px), the mean absolute error of Landsat (via SWORD) estimated widths is lower (Table~\ref{table:river-width-estimation-results}). 
This may be attributed to the additional post-processing steps---such as basin-specific distribution fitting---applied to Landsat products, which can suppress outliers in the width estimates~\cite{allen2018global}.
Nonetheless, looking at median absolute error, a metric less sensitive to outliers, Sentinel widths still result in lower error compared to Landsat.

\noindent\textbf{Combining SWOT and Sentinel predictions might offer a more robust approach for width estimation.} 
Table~\ref{table:river-width-estimation-results} shows that SWOT and Sentinel achieve comparable median absolute error, but Sentinel underperforms in mean absolute error and bias.
This bias stems from the high false positive rate in Sentinel's water masks as detailed in \S~\ref{subsec:expt-water-segmentation}.
Tying Sentinel widths to co-located SWOT measurements might help lessen this overestimation and reduce errors whenever optical observations are degraded by shadows, haze, clouds, or sunglint~\cite{zhu2018automatic}.
Incorporating SWOT's KaRIn point cloud interferometry data---which provides geophysical data and radar echo power of samples---with Sentinel's multispectral data may provide a physically grounded constraint on rivers that can reject optical/spectral outliers~\cite{lin2020global}.
Exploring this direction in future work may yield more reliable river width estimates when the optical data are limited or compromised.

\section{Limitations}
While our dataset presents rivers of varying widths, most rivers come from the Northern Hemisphere, providing an opportunity to expand to rivers in the Southern Hemisphere.
Our width labels also rely on the river reaches defined by SWORD, which tend to simplify multiple or braided channels to a single channel, underestimating the widths in complex sections.
The PlanetScope imagery used in our dataset has high resolution, but it is a commercial product---purchasing requirements limit access to additional data.
Future large-scale applications can use our work to evaluate the trade-off between cost and accuracy.
Additionally, all optical sensors remain vulnerable to haze, sunglint, and cloud cover.
\S~\ref{subsec:limitations} (Appendix) discusses these limitations in detail.

\section{Conclusion}
\label{sec:conclusion}
We introduced RiverScope, a high-resolution dataset bridging machine learning and critical challenges in hydrology.
It provides expertly annotated 3~m/pixel PlanetScope imagery co-registered with public data from SWOT, SWORD, and Sentinel to benchmark different sensor capabilities for monitoring Earth's river systems.

Our benchmark establishes a new state-of-the-art for river width estimation, achieving a median error of 7.2 meters.
This result also highlights a critical trade-off: while our RiverScope dataset is released publicly for research, scaling this high-accuracy approach requires purchasing additional PlanetScope imagery.
This cost must be weighed against the lower performance of freely available Sentinel and Landsat data, which yield larger errors.
We additionally show that using linear adaptors to adapt pretrained models for multispectral data is key to accurate segmentation.

By providing the first high-resolution benchmark for river width---an essential variable for estimating river discharge---RiverScope enables the development of more reliable hydrological models.
This is a significant step towards robust monitoring of global river dynamics. 
We invite the machine learning community to use this resource to advance the state-of-the-art in remote sensing hydrology and create new multi-sensor approaches to characterize our planet's rivers.

\section*{Acknowledgements}
We thank William Tooley for participating in the labeling of images. RD, TS, CG, and SM were supported in part by NASA grant 80NSSC22K1487. Additionally, RD and SM are supported by NSF grant 2329927, CG by NASA grant 80NSSC24K1646, and TR by a contract from the Jet Propulsion Laboratory.

% \clearpage

% \clearpage
\bibliography{aaai2026}

% \clearpage


\appendix
\setcounter{page}{1}
\setcounter{table}{0}   
\renewcommand{\thetable}{A\arabic{table}}
\setcounter{figure}{0}
\renewcommand{\thefigure}{A\arabic{figure}}
\section{Appendix}
\label{sec:appendix}

\subsection{Limitations}
\label{subsec:limitations}
\noindent\textbf{Data distribution. }
Our dataset establishes a strong foundation with rivers of varying widths, primarily focused on the Northern Hemisphere due to the SWOT orbit.
This provides an opportunity to expand the dataset to include more diversity of river systems in the Southern Hemisphere.
The width labels in our work are also based on SWORD. While SWORD effectively defines single-channel reaches, future work could build on our work to address more complex morphologies, such as braided and multi-channel networks, which are challenging in river science.

\noindent\textbf{Cost for large-scale applications.}
To achieve high precision, this study leveraged the high resolution of PlanetScope imagery. Its quality demonstrates the upper bound of what is possible with current technology. For future large-scale applications, our work can be used to evaluate the trade-off between purchasing more PlanetScope data and settling with less accuracy. Overcoming the coarser resolution of Sentinel and Landsat to achieve precise segmentation is a key challenge that our work helps to define. Furthermore, while all optical sensors are susceptible to atmospheric conditions like clouds and haze, the multimodal fusion approaches outlined below offer a robust path to mitigate these effects.

\noindent\textbf{Minimum detectable river width.}
The theoretical minimum detectable width is 3 meters (corresponding to a single pixel). However, based on our experience with previous computer vision models for river detection, we generally consider a river to be “detectable” when it spans at least three visible pixels across its width (i.e., approximately 9 meters wide for RiverScope).

\begin{figure}[t]
    \centering
    \includegraphics[width=0.6\linewidth]{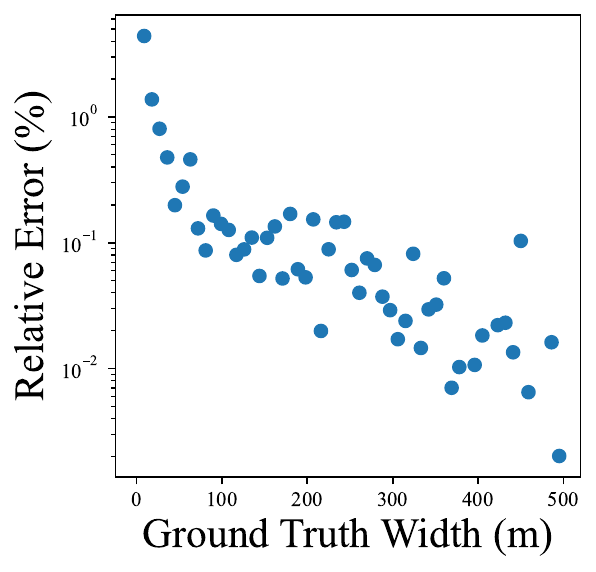}
    \caption{\textbf{Trend of relative error as the ground truth river width decreases}. We generally see higher relative error for smaller rivers, and lower error for larger rivers.
    }
    \label{fig:min-detectable-width-results}
\end{figure}

Analyzing the river width estimation error as a function of the ground truth width, we find that the relative error generally increases with smaller rivers. 
Relative error is defined as $\frac{|y_i - \hat{y}_i|}{y_i} \times 100\%$ for a predicted width $\hat{y}_i$ and corresponding ground truth width $y_i$. 
Figure~\ref{fig:min-detectable-width-results} summarizes this trend by showing the relative error across different ground truth river widths.
The relative error is generally higher for smaller rivers, and lower for larger rivers.
We found that while the model can detect narrow rivers, the relative error decreases significantly to 20\% (9.8 meters MAE) for rivers around 45 meters wide (15 pixels).

\subsection{Future Work}
\textbf{Addressing limitations.} Future research can directly address the limitations through building upon our contributions by (1) exploiting dense temporal stacks of Sentinel/Landsat to refine low-resolution segmentation, and (2) fusing SWOT's precise height measurements with optical imagery. These approaches could effectively reduce false positives in water segmentation and provide a promising direction to explore multimodal learning for river science and water resource management.

\noindent\textbf{Reservoir detection.} Our work primarily focused on river segmentation, though some of our training data can include reservoirs since classify all water pixels. The model can often detect reservoir because of their visual similarity to rivers. However a more detailed evaluation would be needed to assess its accuracy in those cases.

\noindent\textbf{Exploring other pretrained models.} In this paper, we focused on widely used RGB and multispectral pretrained models. However, future work can look into models trained on high-resolution RGB-NIR similar to datasets fMoW~\cite{christie2018functional_fmow} and NAIP~\cite{usda_naip_imagery}. Learning high-resolution features on these datasets could potentially further improve performance.

% While our dataset presents rivers of varying widths, most rivers come from the Northern Hemisphere, leaving other types of rivers that occur in different parts of the world underrepresented.
% Our width labels also rely on the river reaches defined by SWORD, which tend to simplify multiple or braided channels to a single channel, underestimating the widths in complex sections.
% The PlanetScope imagery used in our dataset has high resolution, but it is a commercial product---additional images would need to be purchased, if needed.
% Free optical alternatives such as Sentinel and Landsat provide the same global coverage, but their coarser resolution limit segmentation and width precision.
% Additionally, all optical sensors remain vulnerable to haze, sunglint, and cloud cover.

% Future work can overcome these constraints by (1) exploiting dense temporal stacks of Sentinel/Landsat to refine low-resolution segmentation, and (2) fusing SWOT's precise height measurements with optical imagery.
% These approaches could effectively reduce false positives in water segmentation, and provide a promising direction to explore multi-modal learning for river science and water resource management.

\subsection{Data Labeling Tool}
\label{subsec:supp-data-labeling}
\begin{figure}[t]
    \begin{center}
    \includegraphics[scale=0.5]{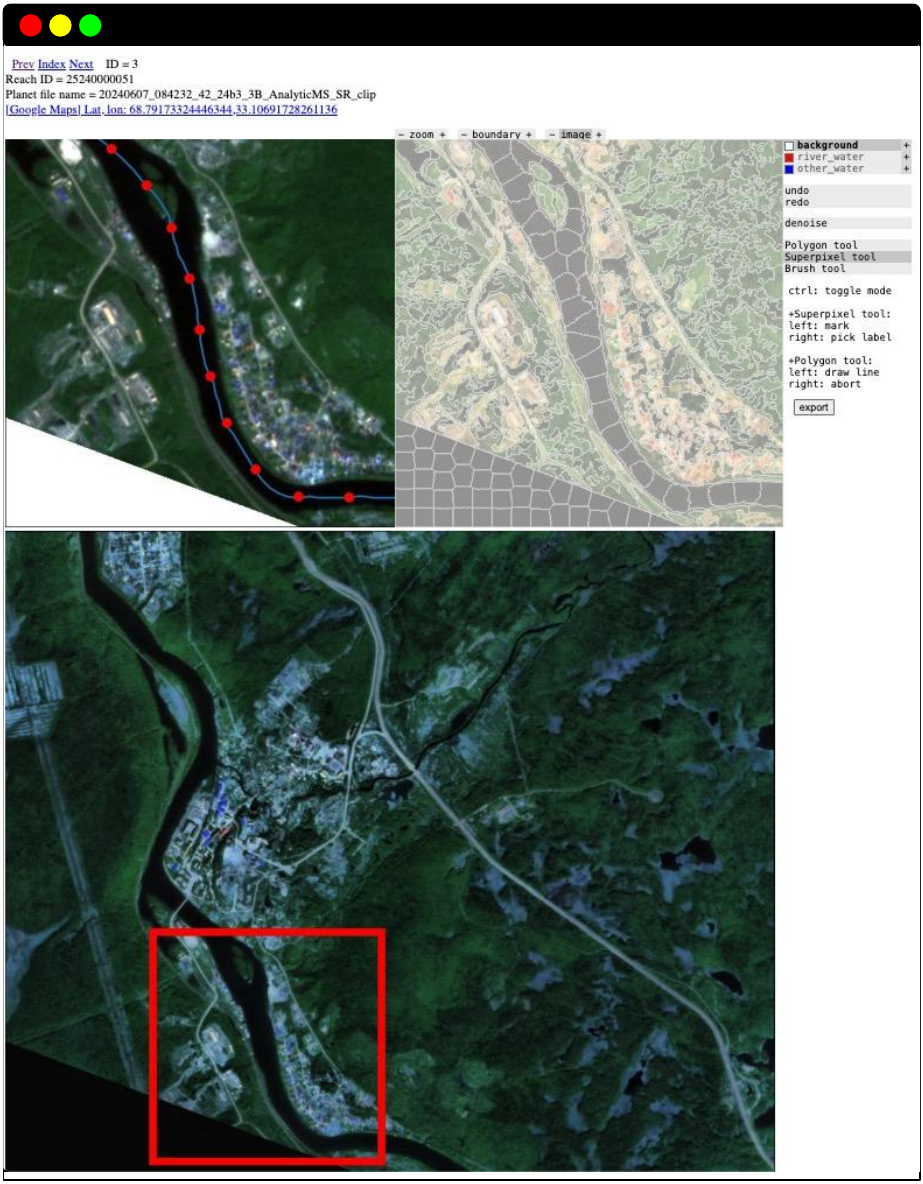}
    \end{center}
    % \vspace{-6pt}
    \caption{\textbf{Data labeling tool visualization}. The tool is a web-based application that all annotators used for labeling PlanetScope images. Various views of the same image are provided to help annotators reliably identify water pixels.}
    \label{fig:supp-data-labeling-tool}
    % \vspace{-5mm}
\end{figure}

Figure~\ref{fig:supp-data-labeling-tool} shows the data annotation tool for labeling PlanetScope images. We use the tool from \citet{tangseng2017looking} as a starting point\footnotemark. The image on the upper left shows the RGB channels of the tile to be labeled, while the image on the upper right shows where annotators can label over the image. The SWORD nodes and reaches are overlaid on the upper left image as additional reference. The transparency and scale of the image on the upper right can be changed to see the finer details. The large image at the bottom shows the source PlanetScope tile, with the red box indicating the current tile being annotated. Additional information such as the reach identifier, latitude, longitude, and Google Maps link are also included in the upper left for the annotators' reference.
\footnotetext{\url{https://github.com/kyamagu/js-segment-annotator}}

Various tools can be used for annotation such as the polygon tool, the brush tool, and the superpixel tool (seen on the toolbar on the upper right). The superpixel tool clusters pixels based on similarity, and annotators can simply select the cluster of pixels to speed up the annotation. The image on the upper right illustrates this, with the superpixels visualized by the polygons. The size of the clusters or superpixels can also be adjusted to be finer or coarser. Annotators can label river water and non-river water by selecting the correct class at the top of the toolbar.

\subsection{Additional RiverScope Details}
\label{subsec:supp-riverscope-details}
The full dataset required for running the RiverScope training, evaluation, and visualization consumes about 10~GB of storage. We include additional details for the data sources and for the model training below.

\noindent\textbf{Northern latitude bias.} The northern latitude bias stems from the prioritization of sites located within the fast sampling orbit of the SWOT satellite, which provides more frequent observations in these regions. This allowed us to maximize the overlap with available SWOT data. Additionally, there is more landmass in the Northern Hemisphere than the Southern Hemisphere. Since our focus is on surface water over land (i.e., rivers), this resulted in a northern-heavy distribution of sites. 

\noindent\textbf{SWORD and SWOT details.} The SWORD dataset used was obtained from SWORDv16.
SWOT data on pixel clouds (PIXC), nodes (RiverSP Node), and reaches (RiverSP Reach) were obtained from the composite release identifier PIC0, version 1.0 of the product (downloaded May 2, 2025).
To filter the corresponding SWOT data (since there can be multiple SWOT matches for a given PlanetScope area and timestamp), we select the SWOT sample that has the largest intersection with the PlanetScope image.
In particular, we choose the pixel cloud with the most number of pixel cloud points within the PlanetScope image, and the node/reach aggregated data that had the most number of nodes within the PlanetScope image.
For reference, we also include all corresponding SWOT and SWORD IDs in the dataset to download the raw data.

\noindent\textbf{PlanetScope details.} The PlanetScope data released as part of RiverScope were normalized using min-max normalization per image. That is, for each image, we get the maximum and the minimum values for normalization. Pixels with no data were filled with 0 values. We don't use the minimum/maximum across the whole dataset since the raw data values are proprietary. The raw data can be purchased directly from PlanetLabs~\cite{planetlabs} using the PlanetScope image IDs released in RiverScope. %Additionally, while most of the collected data are from the Northern Hemisphere, this could be attributed to the larger land area compared to the Southern Hemisphere.

\begin{table*}[t]
    \setlength{\tabcolsep}{5pt} 
    \small
    \begin{center}
     
    \begin{tabular}{l l |c c c }
        \toprule
	Segmentation Model & Backbone & Drop &  Linear  & Random	\\
	\midrule
	\multirow{4}{*}{DeepLabv3} & MV2 & 91.35 $\pm$ 1.18 & 93.73 $\pm$ 0.38 & 93.57 $\pm$ 0.54\\
	  & RN50 & 92.42 $\pm$ 0.47 & 93.87 $\pm$ 0.39 & 90.33 $\pm$ 1.91\\
	  & Swin-B & 86.27 $\pm$ 0.34 & 86.64 $\pm$ 0.66 & 81.83 $\pm$ 0.74\\
	  & Swin-T & 85.37 $\pm$ 0.86 & 85.79 $\pm$ 0.96 & 81.24 $\pm$ 0.81\\
        \midrule
	\multirow{2}{*}{DPT} & ViT-B/16 & 91.02 $\pm$ 1.10 & 91.20 $\pm$ 1.05 & 92.01 $\pm$ 0.52\\
	  & ViT-L/16 & 90.91 $\pm$ 1.63 & 93.07 $\pm$ 0.64 & 91.05 $\pm$ 1.09\\
        \midrule
	\multirow{4}{*}{FPN}  & MV2 & 91.06 $\pm$ 1.22 & 93.38 $\pm$ 0.66 & 93.70 $\pm$ 0.44\\
	  & RN50 & 91.37 $\pm$ 0.79 & 93.48 $\pm$ 0.71 & 91.98 $\pm$ 1.36\\
	  & Swin-B & 92.51 $\pm$ 0.45 & 92.33 $\pm$ 2.43 & 90.57 $\pm$ 0.93\\
	  & Swin-T & 92.44 $\pm$ 0.21 & 93.27 $\pm$ 0.48 & 90.93 $\pm$ 1.65\\
        \midrule
	\multirow{4}{*}{UNet} & MV2 & 92.20 $\pm$ 0.98 & 94.01 $\pm$ 1.01 & 92.17 $\pm$ 1.09\\
	  & RN50 & 92.29 $\pm$ 1.09 & 94.16 $\pm$ 0.37 & 92.24 $\pm$ 0.82\\
	  & Swin-B & 93.13 $\pm$ 0.71 & 94.12 $\pm$ 0.55 & 92.42 $\pm$ 1.06\\
	  & Swin-T & 93.35 $\pm$ 0.53 & 93.42 $\pm$ 1.02 & 92.68 $\pm$ 0.76\\
        \midrule
        \midrule
        Average    &   & 91.12 $\pm$ 2.45 & \textbf{92.32 $\pm$ 2.76} & 90.48 $\pm$ 3.92 \\
    
     \bottomrule
     \end{tabular}
     \caption{\textbf{F1 score of different ways of adapting a 4-channel image to various models}. The average F1 score and the standard deviations are indicated for the different configurations. The numbers are computed from five runs for each of the model configurations.}
     \label{table:supp-adaptor-quantitative}
     \end{center}
\end{table*}

\begin{table*}[t]
    \setlength{\tabcolsep}{2pt}    
    \small
    \begin{center}
     
    \begin{tabular}{llccc}
    \toprule
    Segmentation Model & Backbone & RGBR & RGB+Random & Linear \\
    \midrule
    \multirow{2}{*}{DeepLabv3} & MV2 & $93.6 \pm 1.0$ & $93.1 \pm 1.1$ & $93.7 \pm 0.4$ \\
    & RN50 & $93.8 \pm 0.2$ & $93.5 \pm 0.4$ & $93.9 \pm 0.4$ \\
    \midrule
    \multirow{2}{*}{FPN} & MV2 & $94.1 \pm 0.1$ & $93.7 \pm 0.5$ & $93.4 \pm 0.7$ \\
    & RN50 & $93.1 \pm 0.8$ & $92.4 \pm 1.3$ & $93.5 \pm 0.7$ \\
    \midrule
    \multirow{2}{*}{UNet} & MV2 & $92.8 \pm 1.4$ & $93.3 \pm 0.4$ & $94.0 \pm 1.0$ \\
    & RN50 & $93.9 \pm 0.3$ & $92.8 \pm 0.4$ & $94.2 \pm 0.4$ \\
    \midrule
    \midrule
    \multicolumn{2}{l}{Average} & $93.6 \pm 0.8$ & $93.1 \pm 0.8$ & \textbf{93.8 $\pm$ 0.6} \\
    \bottomrule
    \end{tabular}
     \caption{\textbf{Additional ways to handle NIR channel for pretrained RGB models}. RGB+Random and RGBR use the original weights for the RGB channels in the first layer; they only differ in the way the NIR channel is handled. RGB+Random uses randomly initialized weights for NIR, while RGBR reuses the weights from the red channel for NIR. Linear still results in overall better performance.}
     \label{table:additional-nir-adaptor-results}
    \end{center}
\end{table*}

\noindent\textbf{RiverScope models training.} All RiverScope models were trained for about an hour per model on an NVIDIA L40S GPU for 50 epochs, and a batch size of 32.
A checkpoint is saved after every epoch, and the best model among all epochs is selected based on the validation set F1-score.
For large models using ViT-L/16 and Swin-B, to speed up training, we use 2 NVIDIA L40S GPUs.
The optimal learning rates found for each of the model configurations will be made available in the code repository.

\begin{table}[t]
    \setlength{\tabcolsep}{3pt}    
    \small
    \begin{center}
    \begin{tabular}{l | r r r }
    \toprule
    Model & F1 Score & Precision & Recall \\
    \midrule
    PlanetScope NDWI & 72.77\% & 60.55\% & 91.18\% \\
    PlanetScope CNN & 78.77\% & 81.09\% & 76.59\% \\
    \bottomrule
    \end{tabular}
     \caption{\textbf{River segmentation quantitative results of PlanetScope baselines}. }
     \label{table:supp-river-segmentation-results-planet-baselines}
    \end{center}
\end{table}

\subsection{Sentinel Models}
\label{subsec:supp-sentinel-model}
\begin{figure}[t]
    \begin{center}
    \includegraphics[width=\linewidth]{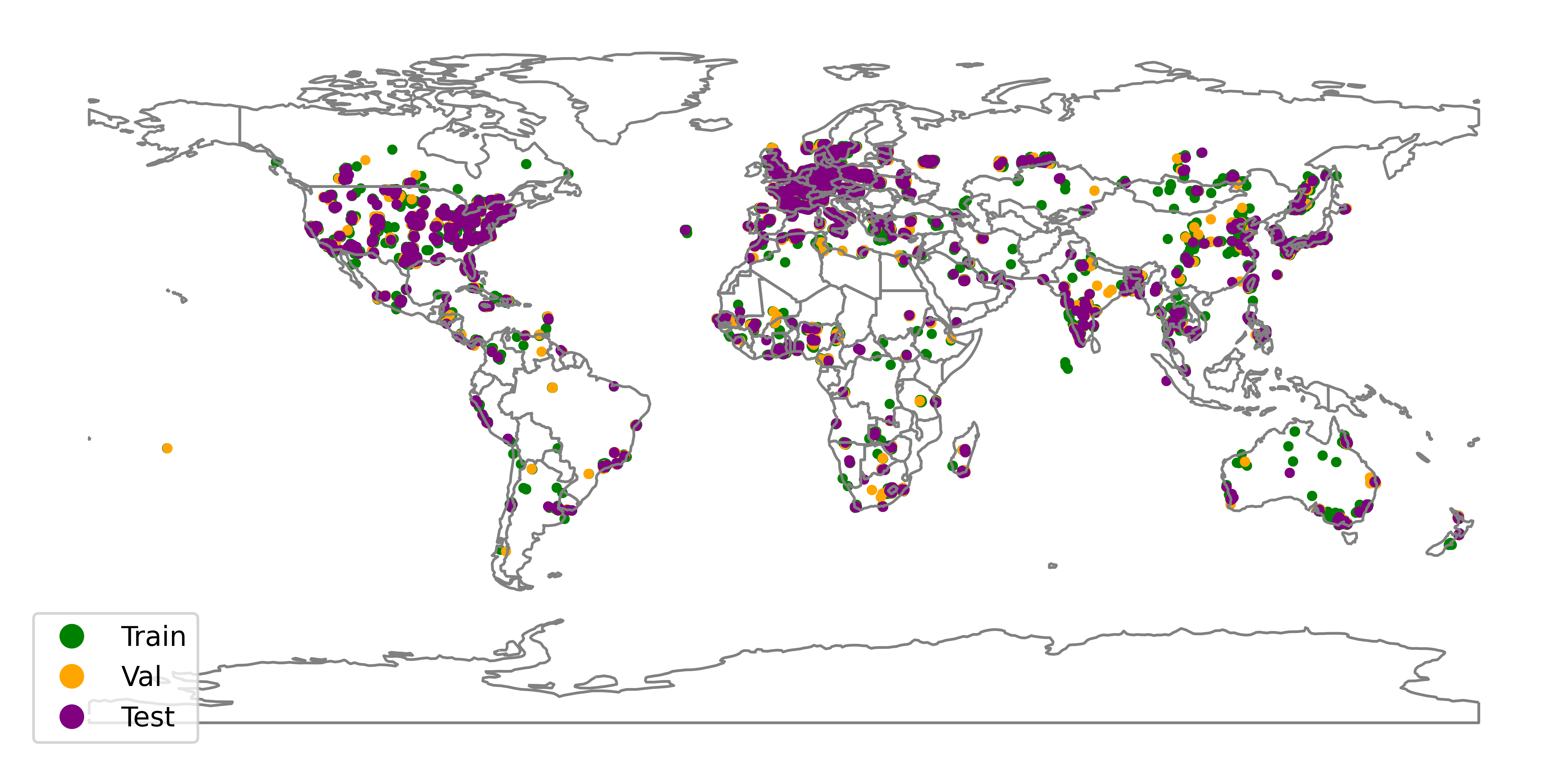}
    \end{center}
    % \vspace{-6pt}
    \caption{\textbf{Data distribution of data used to train and validate Sentinel models}. Water segmentation labels are taken from WorldCover with a resolution of 10~m/pixel.}
    \label{fig:sentinel-data-distribution}
    % \vspace{-5mm}
\end{figure}

We use labeled data from WorldCover in 2020-2021 which has segmentation labels for land cover, including water segmentation labels. Only water labels are used for training Sentinel models. The training data was uniformly sampled on a global-scale, and is adopted from SatlasPretrain~\cite{bastani2023satlaspretrain}. Figure~\ref{fig:sentinel-data-distribution} shows the distribution of the data with corresponding labels from WorldCover.

Sentinel trained models use 6 bands for training: red, blue, green, NIR, SWIR1, and SWIR2 (bands 2,3,4,8,11,12). This follows previous work~\cite{daroya2025improving, isikdogan2019seeing} that predict masks on satellite images. Similar to RiverScope trained models, we use the Adam optimizer to train all models with the learning rate selected based on the performance on the validation set. The same binary cross entropy loss was also used. Segmentation performance was evaluated on the RiverScope test set, with the Sentinel predictions reprojected and upsampled to match the view and the resolution of the labels. Training time is about 12 hours on an NVIDIA L40S GPU for 50 epochs with a batch size of 32.

\subsection{Additional Results}
\label{subsec:supp-addtl-results}

\subsubsection{Adaptors}
Table~\ref{table:supp-adaptor-quantitative} shows quantitative results of different adaptors across different models. We find that using a linear adaptor (Linear) on pretrained models that only use RGB as inputs results in better overall performance than simply dropping the extra channel (Drop), or training with all 4 channels from scratch (Random). The average and standard deviations are computed from multiple runs for each backbone and pretraining setup.

We also explore other ways to handle the NIR channel for pretrained RGB models such as RGB+Random and RGBR.
RGB+Random keeps the original RGB pre-trained weights for the first layer of the model, and adds an additional channel with randomly initialized weights to handle the NIR channel.
RGBR is similar to RGB+Random, but instead of randomly initializing the weights for the additional channel, the weights for the first channel are used.
Scaling is applied to both methods to ensure the magnitude of the features do not change when fed to subsequent layers.
Table~\ref{table:additional-nir-adaptor-results} shows the results, where using a linear adaptor still resulted in overall better performance.

\begin{table*}[t]
    \setlength{\tabcolsep}{2pt}    
    \small
    \begin{center}
     
    \begin{tabular}{l l l | r r r | r r r}
    \toprule
    Segmentation & \multirow{2}{*}{Backbone} & \multirow{2}{*}{Pretraining} & \multicolumn{3}{|c|}{RiverScope} & \multicolumn{3}{|c}{Sentinel} \\
    Model &  & & F1 & Prec & Rec & F1 & Prec & Rec \\
    \midrule
    \multirow{6}{*}{DeepLabv3} & MV2 & Supervised (ImageNet1k) & 93.59 & 94.40 & 92.86 & 42.67 & 55.96 & 42.35 \\
      & RN50 & Supervised (ImageNet1k) & 93.79 & 92.58 & 95.05 & 56.17 & 42.25 & 83.78 \\
      & RN50 & MoCov3 (ImagetNet1k) & 93.92 & 93.64 & 94.20 & 58.16 & 42.80 & 90.90 \\
      & RN50 & SeCo & 92.96 & 92.49 & 93.46 & 56.97 & 42.24 & 87.63 \\
      & Swin-B & Supervised (ImageNet1k) & 86.89 & 86.53 & 87.26 & 54.07 & 39.60 & 85.19 \\
      & Swin-T & Supervised (ImageNet1k) & 85.81 & 85.23 & 86.42 & 53.90 & 39.66 & 84.15 \\
    \midrule
    \multirow{6}{*}{DPT} & ViT-B/16 & Supervised (ImageNet1k) & 91.21 & 89.50 & 92.99 & 60.48 & 44.62 & 93.81 \\
      & ViT-B/16 & CLIP & 93.30 & 92.29 & 94.36 & 60.76 & 44.96 & 93.68 \\
      & ViT-B/16 & DINO & 91.62 & 89.18 & 94.29 & 60.06 & 44.24 & 93.50 \\
      & ViT-B/16 & MoCov3 (ImagetNet1k) & 90.80 & 88.43 & 93.50 & 60.23 & 44.61 & 92.71 \\
      & ViT-B/16 & Prithvi & 90.51 & 90.64 & 90.49 & 60.15 & 44.73 & 91.79 \\
      & ViT-L/16 & Supervised (ImageNet1k) & 92.88 & 90.01 & 95.94 & 60.85 & 45.02 & 93.89 \\
    \midrule
    \multirow{9}{*}{FPN} & MV2 & Supervised (ImageNet1k) & 93.26 & 91.38 & 95.23 & 57.78 & 42.79 & 89.11 \\
      & RN50 & Supervised (ImageNet1k) & 93.64 & 93.15 & 94.14 & 58.19 & 42.31 & 93.31 \\
      & RN50 & MoCov3 (ImagetNet1k) & 93.39 & 91.49 & 95.50 & 59.11 & 43.44 & 92.56 \\
      & RN50 & SeCo & 93.55 & 91.27 & \textbf{96.03} & 60.85 & 57.33 & 79.16 \\
      & RN50 & SatlasNet & 93.41 & 91.24 & 95.72 & 56.50 & 42.29 & 85.30 \\
      & Swin-B & SatlasNet & 93.60 & 91.88 & 95.40 & \textbf{67.11} & \textbf{58.64} & 87.77 \\
      & Swin-T & SatlasNet & 94.35 & 93.34 & 95.39 & 63.21 & 57.74 & 80.65 \\
      & Swin-B & Supervised (ImageNet1k) & 93.38 & 91.61 & 95.25 & 60.49 & 44.68 & 93.61 \\
      & Swin-T & Supervised (ImageNet1k) & 93.27 & 91.84 & 94.75 & 60.89 & 44.89 & 94.61 \\
    \midrule
    \multirow{6}{*}{UNet} & MV2 & Supervised (ImageNet1k) & 93.84 & 91.80 & \textbf{96.03} & 59.27 & 43.49 & 93.08 \\
      & RN50 & Supervised (ImageNet1k) & 94.05 & 93.33 & 94.80 & 57.44 & 41.74 & 92.43 \\
      & RN50 & MoCov3 (ImagetNet1k) & 94.20 & 93.62 & 94.80 & 60.59 & 44.58 & 94.54 \\
      & RN50 & SeCo & \textbf{94.39} & \textbf{94.21} & 94.62 & 55.74 & 44.22 & 79.02 \\
      & Swin-B & Supervised (ImageNet1k) & 94.09 & 93.01 & 95.22 & 60.92 & 45.00 & 94.27 \\
      & Swin-T & Supervised (ImageNet1k) & 93.30 & 91.89 & 94.84 & 61.09 & 45.09 & \textbf{94.74} \\

    \bottomrule
    \end{tabular}
     \caption{\textbf{River segmentation quantitative results}. F1 score (F1), precision (Prec), and recall (Rec) are reported for RiverScope and Sentinel trained models.}
     \label{table:river-segmentation-results}
    \end{center}
\end{table*}

\subsubsection{Water Segmentation Results}
\begin{figure*}[t]
    \begin{center}
    \includegraphics[width=0.6\linewidth]{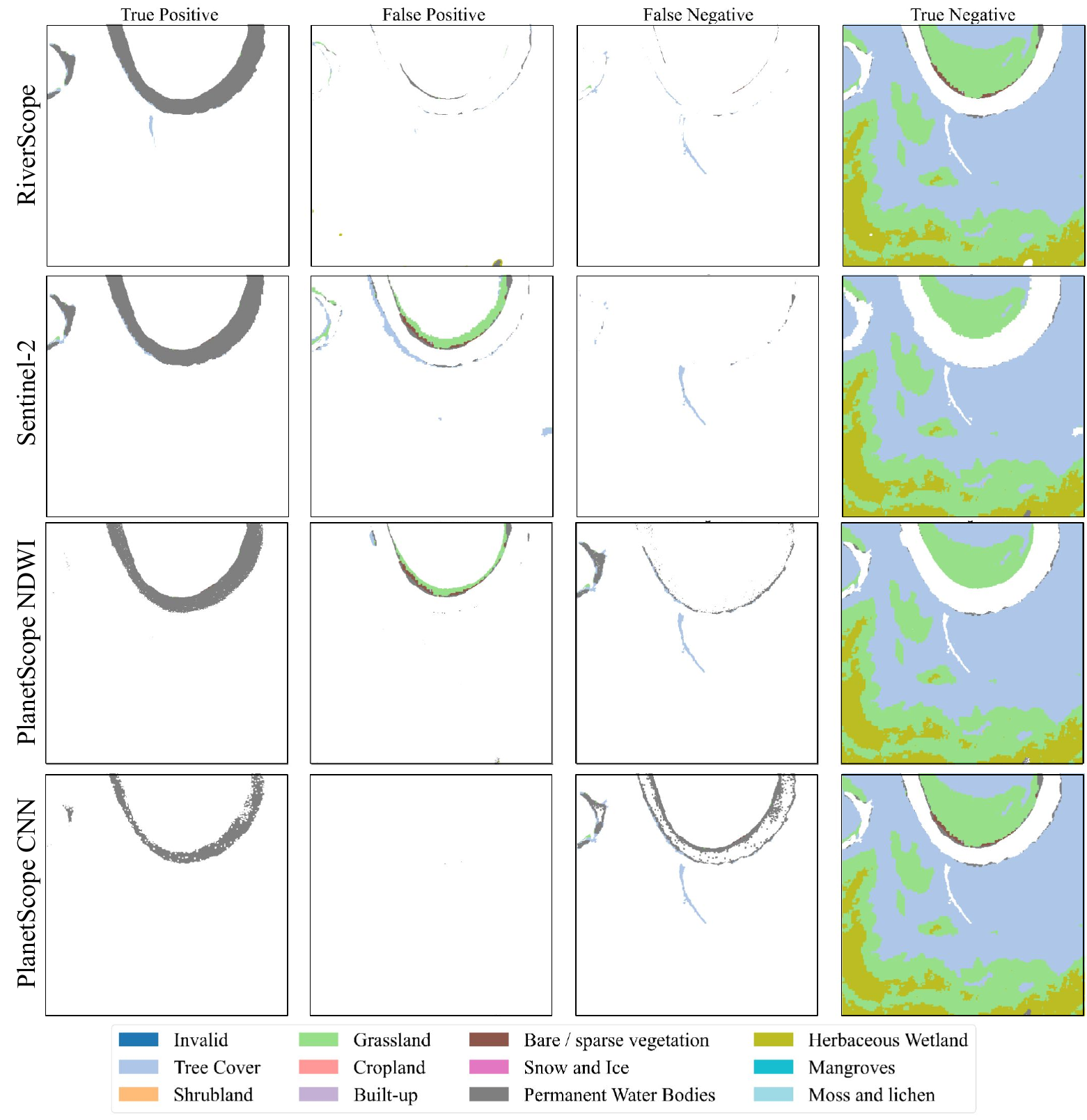}
    \end{center}
    % \vspace{-6pt}
    \caption{\textbf{Segmentation error visualization}. Predictions from the best performing RiverScope and Sentinel models and baseline models are visualized above, with sources of error visualized using land use and land cover labels from WorldCover. Errors come from both the false positives and false negatives. For all models, much of the error comes from misclassifying `Tree cover' and `Grassland' as water. Additionally, we find Sentinel and PlanetScope NDWI models have less precision around the boundaries of the water body, resulting in larger false positive errors. In contrast, PlanetScope CNN tends to underpredict water pixels with more false negatives.}
    \label{fig:supp-error-visualization}
    % \vspace{-5mm}
\end{figure*}

In addition to including quantitative results (Table~\ref{table:river-segmentation-results}, Table~\ref{table:supp-river-segmentation-results-planet-baselines}), we also include results from additional error analyses (Table~\ref{table:supp-false-positive-seg}, Figure~\ref{fig:supp-error-visualization}).
All quantitative results are averaged over five runs.
To evaluate and analyze the remaining sources of error for models, we use existing land use and land cover (LULC) labels from WorldCover.
For all false positive and false negative pixels, we find the corresponding LULC labels to find out the likely class that caused misclassifications.
While the LULC labels have a lower 10m/pixel resolution, these nonetheless give a good sense of the sources of error from the models.

\noindent\textbf{Quantitative RiverScope and Sentinel water segmentation results.} Table~\ref{table:river-segmentation-results} shows the raw numbers that compare the segmentation performance of RiverScope trained models and Sentinel trained models. We include the F1 score, precision, and recall. We observe that RiverScope trained models have superior F1 score performance for all models, and has particularly higher precision. Sentinel trained models tend to have high recall, indicating the ability to find actual water pixels, but struggle with precision. This ultimately led to the lower performance of Sentinel models, since the overall F1 score is pulled down by the lower precision.

\noindent\textbf{PlanetScope NDWI tends to overestimate water boundaries and is sensitive to manmade structures.} Table~\ref{table:supp-river-segmentation-results-planet-baselines} shows the raw numbers for PlanetScope baselines CNN and NDWI. 
Despite the higher resolution satellite imagery, NDWI segments beyond the water boundaries, resulting in very precision.
In addition, unlike other models that distinguish water from manmade structures (e.g., buildings), NDWI is sensitive to these classes. Table~\ref{table:supp-false-positive-seg} shows that only NDWI has a non-zero false positive rate on `Built-up', consistent with findings from \citet{xu2006modification}.

\noindent\textbf{PlanetScope CNN misses a significant number of water pixels.} 
Despite training on high-resolution data, the CNN baseline tends to miss a lot of water pixels. 
Table~\ref{table:supp-river-segmentation-results-planet-baselines} shows CNN achieves a significantly lower recall than NDWI.
Figure~\ref{fig:supp-error-visualization}, Figure~\ref{fig:supp-visualization-results} also show the same behavior of missing water pixels and having a large number of false negatives.
This is likely due to the absence of pretraining which limits its ability to capture subtler, and possibly more complex, features that characterize water.

\noindent\textbf{Finer spatial detail helps improve performance of RiverScope trained models.} 
Figure~\ref{fig:supp-error-visualization} shows sample visualizations of the correctly and incorrectly classified pixels. 
Correctly classified pixels appear as true positive and true negative pixels, while incorrectly classified pixels appear as false positive and false negative pixels.
Looking at the last column visualizing true negatives (i.e., pixels that were correctly identified as non-water), RiverScope trained models were able to identify `Bare/sparse vegetation' along the river bank as non-water likely due to the finer spatial detail RiverScope provides. In contrast, the Sentinel trained model was not able to do the same. This is consistent with the results in Table~\ref{table:supp-false-positive-seg}.

% \noindent\textbf{`Tree cover' and `Grassland' are frequently mistaken as water pixels for both RiverScope and Sentinel trained models.} Using land use and land cover labels, we identify the likely cause of the misclassification of pixels. Table~\ref{table:supp-false-positive-seg} shows the breakdown of false positive pixels (i.e., pixels that were classified as water by the models, but are not water based on ground truth labels). `Tree cover' and `Grassland' have the higher percentage among false positive pixels. `Herbaceous Wetland' is another source of confusion for the models, likely since these typically occur along the edges of lakes and rivers. 

\begin{table*}[t]
    \small
    \begin{center}
    \begin{tabular}{l |r r r r}
    \toprule
    & RiverScope & Sentinel & PlanetScope CNN & PlanetScope NDWI \\
    \midrule
    Tree Cover & 33.66\% & 25.44\% & 25.02\% & 41.00\% \\
    Shrubland & 1.15\% & 0.97\% & 0.19\% & 0.79\% \\
    Grassland & 19.92\% & 63.73\% & 13.59\% & 21.46\% \\
    Cropland & 0.00\% & 0.00\% & 0.00\% & 0.05\% \\
    Built-up & 0.00\% & 0.00\% & 0.00\% & 0.12\% \\
    Bare / sparse vegetation & 11.64\% & 1.91\% & 22.04\% & 15.12\% \\
    Snow and Ice & 0.05\% & 0.00\% & 0.00\% & 0.02\% \\
    Herbaceous Wetland & 16.15\% & 3.93\% & 22.71\% & 9.82\% \\
    Mangroves & 0.00\% & 0.00\% & 0.00\% & 0.00\% \\
    Moss and lichen & 12.37\% & 3.43\% & 10.30\% & 10.23\% \\
    Invalid / no data & 5.06\% & 0.57\% & 6.15\% & 1.38\% \\
     \bottomrule
     \end{tabular}
     \caption{\textbf{Sources of false positives on water segmentation}. Using WorldCover land use and land cover labels, we determine the likely sources of error for the different models. A sample visualization is available in Figure~\ref{fig:supp-error-visualization}.}
     \label{table:supp-false-positive-seg}
     \end{center}
\end{table*}

\subsubsection{River Width Estimation Results}
\label{subsubsec:supp-river-width-estimate-results}

\noindent\textbf{Quantitative river width estimation performance.} 
Table~\ref{table:addtl-river-width-estimation-results} shows the raw numbers from estimating river widths (median absolute error), while Figure~\ref{fig:scatter-river-width-estimation-results} shows the distribution of the predictions with respect to the ground truth. 
RiverScope derived widths consistently perform better than baselines, with as much as a 400~m improvement compared to corresponding Sentinel derived widths (Table~\ref{table:addtl-river-width-estimation-results}).
We also show that RiverScope estimates track the $y=x$ line more closely in Figure~\ref{fig:scatter-river-width-estimation-results}, indicating that RiverScope predictions are more closely aligned with the ground truth widths.
Sentinel and SWOT tend to estimate widths to be larger than ground truth, while Landsat estimates tend to estimate widths to be smaller.

\begin{figure*}[t]
    \begin{center}
    \includegraphics[width=0.8\linewidth]{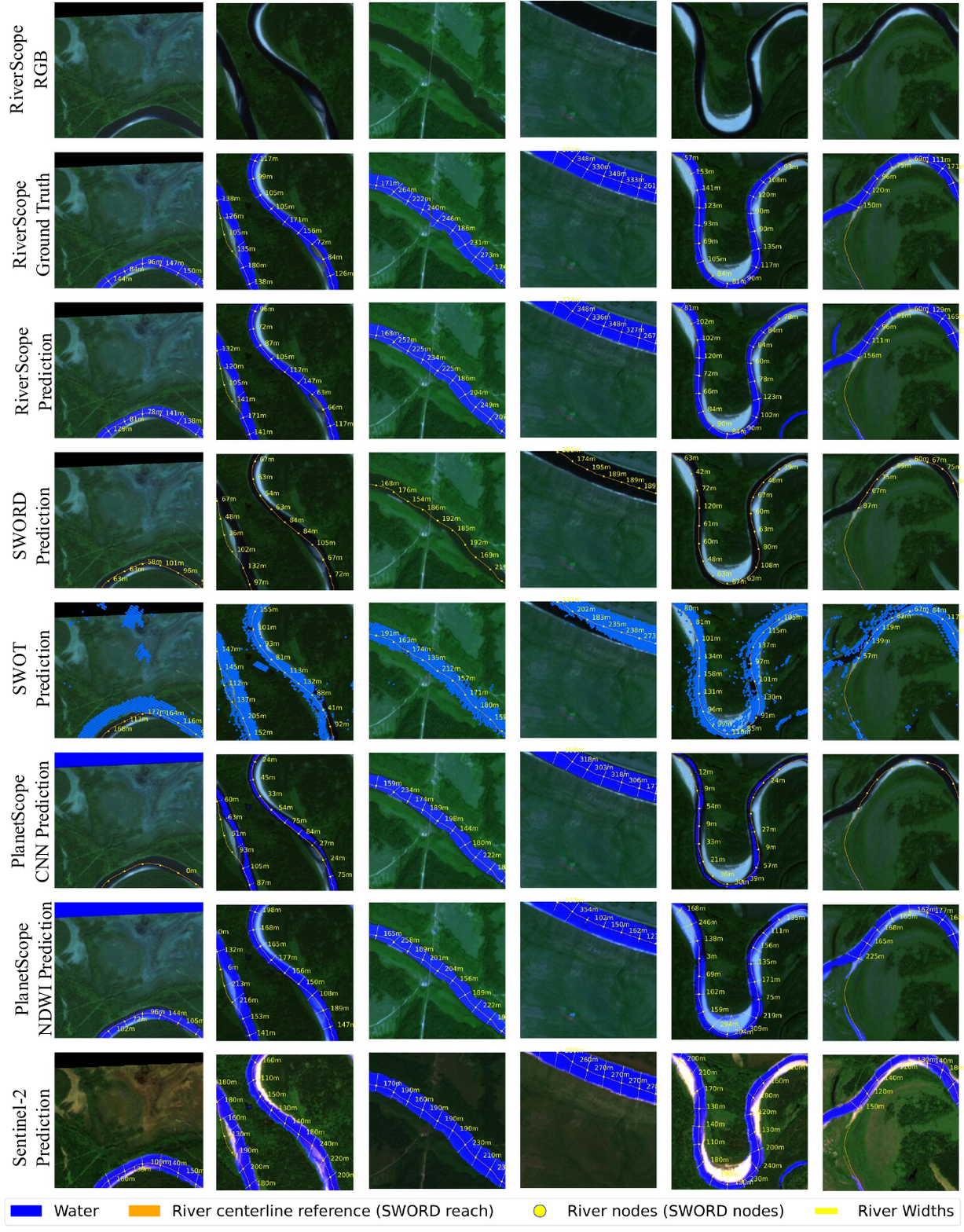}
    \end{center}
    % \vspace{-6pt}
    \caption{\textbf{Segmentation and river width estimation visualization}. }
    \label{fig:supp-visualization-results}
    % \vspace{-5mm}
\end{figure*}

\noindent\textbf{Randomly initialized high-resolution model performs better than pretrained Sentinel model.} Table~\ref{table:river-width-estimation-results} shows PlanetScope CNN has overall better performance than the best performing Sentinel and Landsat models. This highlights the advantage of using higher resolution data over low-resolution data, even without pretraining on large-scale datasets. However, PlanetScope CNN is still prone to significant bias and error compared to pretrained models with RiverScope. This can be attributed to the CNN model missing majority of the water pixels (see Figure~\ref{fig:supp-visualization-results}).

\noindent\textbf{RN50 pretrained on ImageNet has competitive performance.} Although the task requires reasoning on satellite images to discrimate between water and non-water pixels, our results show that pretraining on ImageNet1k (supervised) still results in superior performance. This is consistent with results from recent work that also show pretraining on ImageNet1k is still competitive compared to self-supervised remote sensing methods~\cite{lahrichi2025self}.

\noindent\textbf{FPN and UNet segmentation models are reliable.} For both RiverScope and Sentinel trained models, backbones and pretraining methods that use FPN and UNet generally have better performance in river segmentation and width estimation. Their multiscale encoders, reinforced by rich skip connections---dense, stage-by-stage concatenations in UNet and lateral top-down fusion in FPN---preserve fine-grained edge detail while supplying high-level context. This combination lets the network separate water from spectrally similar land classes more effectively, leading to better water masks and, consequently, more accurate width estimates.

\noindent\textbf{BCE loss results in best performance.} Table~\ref{tab:loss-fns-explore} shows the performance of the different settings using different training losses. We explored several alternative loss functions commonly used in binary segmentation, including adaptive max-pooling~\cite{isikdogan2019seeing}, Dice loss~\cite{roy2017relaynet}, and centerline-aware loss~\cite{shit2021cldice}. However, in our experiments across multiple architectures and pre-training strategies, binary cross-entropy (BCE) consistently yielded the best average performance. While alternative losses can offer improvements in specific scenarios, BCE was the most robust and generalizable choice for our dataset and task.

\noindent\textbf{Additional qualitative results.} Figure~\ref{fig:supp-visualization-results} displays the river segmentation and width estimation results of different baselines. Width estimates are annotated beside each indicated node. Some models fail to segment water and effectively have an estimated width of 0~m.
The results here are consistent with previous observations made.
PlanetScope CNN's tendency to miss water pixels is apparent.
In addition, Sentinel and PlanetScope NDWI are less precise around boundaries where features like sandbars are mistakenly classified as water.

\begin{table*}[t]
    \setlength{\tabcolsep}{3pt}
    \small
    \centering
    \begin{tabular}{l l l | r r r }
    \toprule
    Segmentation model & Backbone & Pretraining & RiverScope & Sentinel & Improvement \\
    \midrule
    \multirow{6}{*}{DeepLabv3} & MV2 & Supervised (ImageNet1k) & 12.0 & 123.0 & 111.0 \\
      & RN50 & Supervised (ImageNet1k) & 9.0 & 79.4 & 70.4 \\
      & RN50 & MoCov3 (ImagetNet1k) & 9.0 & 123.4 & 114.4 \\
      & RN50 & SeCo & 12.0 & 126.0 & 114.0 \\
      & Swin-B & Supervised (ImageNet1k) & 40.5 & 310.8 & 270.3 \\
      & Swin-T & Supervised (ImageNet1k) & 34.7 & 384.7 & 350.0 \\
    \midrule
    \multirow{6}{*}{DPT} & ViT-B/16 & Supervised (ImageNet1k) & 19.5 & 88.0 & 68.5 \\
      & ViT-B/16 & CLIP & 10.5 & 75.0 & 64.5 \\
      & ViT-B/16 & DINO & 13.0 & 91.3 & 78.3 \\
      & ViT-B/16 & MoCov3 (ImageNet1k) & 22.5 & 86.8 & 64.3 \\
      & ViT-B/16 & Prithvi & 15.0 & 76.0 & 61.0 \\
      & ViT-L/16 & Supervised (ImageNet1k) & 16.3 & 61.4 & 45.1 \\
    \midrule
    \multirow{8}{*}{FPN} & MV2 & Supervised (ImageNet1k) & 11.3 & 69.3 & 57.9 \\
      & RN50 & Supervised (ImageNet1k) & \textbf{7.2} & 103.0 & 95.8 \\
      & RN50 & MoCov3 (ImagetNet1k) & 12.0 & 88.0 & 76.0 \\
      & RN50 & SeCo & 8.6 & 54.0 & 45.4 \\
      & RN50 & SatlasNet & 9.0 & 59.5 & 50.5 \\
      & Swin-B & SatlasNet & 11.5 & \textbf{39.0} & 27.5 \\
      & Swin-T & SatlasNet & 10.9 & 56.4 & 45.5 \\
      & Swin-B & Supervised (ImageNet1k) & 12.0 & 68.4 & 56.4 \\
      & Swin-T & Supervised (ImageNet1k) & 9.0 & 71.0 & 62.0 \\
    \midrule
    \multirow{6}{*}{UNet} & MV2 & Supervised (ImageNet1k) & 10.0 & 120.5 & 110.5 \\
      & RN50 & Supervised (ImageNet1k) & 9.0 & 437.6 & 428.6 \\
      & RN50 & MoCov3 (ImagetNet1k) & 8.4 & 71.5 & 63.1 \\
      & RN50 & SeCo & 7.5 & 109.2 & 102.0 \\
      & Swin-B & Supervised (ImageNet1k) & 9.0 & 57.4 & 48.4 \\
      & Swin-T & Supervised (ImageNet1k) & 9.0 & 60.3 & 51.3 \\

    \bottomrule
    \end{tabular}
     \caption{\textbf{River width estimation results}. Median absolute error (m) is reported for RiverScope and Sentinel trained models.}
     \label{table:addtl-river-width-estimation-results}
\end{table*}

\begin{table*}[]
    \setlength{\tabcolsep}{3pt}
    \small
    \centering
\begin{tabular}{lll | rrrr}
    \toprule
\textbf{Segmentation Model} & \textbf{Backbone} & \textbf{Pre-training}   & \multicolumn{1}{l}{\textbf{adaptive maxpool}} & \multicolumn{1}{l}{\textbf{dice}} & \multicolumn{1}{l}{\textbf{bce+cldice}} & \multicolumn{1}{l}{\textbf{bce}} \\
    \midrule
\multirow{6}{*}{DeepLabv3}                   & MV2               & Supervised (ImageNet1k) & 93.2\%                                        & 94.3\%                            & 92.3\%                                  & 93.6\%                           \\
                   & RN50              & Supervised (ImageNet1k) & 92.8\%                                        & 93.3\%                            & 91.9\%                                  & 93.8\%                           \\
                   & RN50              & MoCov3 (ImagetNet1k)    & 90.9\%                                        & 94.4\%                            & 93.0\%                                  & 93.9\%                           \\
                   & RN50              & SeCo                    & 92.8\%                                        & 93.5\%                            & 92.9\%                                  & 93.0\%                           \\
                   & Swin-B            & Supervised (ImageNet1k) & 83.5\%                                        & 87.3\%                            & 79.5\%                                  & 86.9\%                           \\
                   & Swin-T            & Supervised (ImageNet1k) & 85.5\%                                        & 85.9\%                            & 75.1\%                                  & 85.8\%                           \\
    \midrule
\multirow{6}{*}{DPT}                         & ViT-B/16          & Supervised (ImageNet1k) & 87.0\%                                        & 89.8\%                            & 87.7\%                                  & 91.2\%                           \\
                         & ViT-B/16          & CLIP                    & 93.4\%                                        & 92.8\%                            & 92.2\%                                  & 93.3\%                           \\
                         & ViT-B/16          & DINO                    & 90.8\%                                        & 90.6\%                            & 91.8\%                                  & 91.6\%                           \\
                         & ViT-B/16          & MoCov3 (ImagetNet1k)    & 91.1\%                                        & 86.8\%                            & 91.6\%                                  & 90.8\%                           \\
                         & ViT-B/16          & Prithvi                 & 90.6\%                                        & 90.1\%                            & 91.1\%                                  & 90.5\%                           \\
                         & ViT-L/16          & Supervised (ImageNet1k) & 93.1\%                                        & 87.4\%                            & 93.4\%                                  & 92.9\%                           \\
    \midrule
\multirow{9}{*}{FPN}                         & MV2               & Supervised (ImageNet1k) & 93.2\%                                        & 94.3\%                            & 93.3\%                                  & 93.3\%                           \\
                         & RN50              & Supervised (ImageNet1k) & 93.0\%                                        & 94.3\%                            & 93.0\%                                  & 93.6\%                           \\
                         & RN50              & MoCov3 (ImagetNet1k)    & 92.9\%                                        & 93.6\%                            & 93.7\%                                  & 93.4\%                           \\
                         & RN50              & SeCo                    & 94.0\%                                        & 94.1\%                            & 89.5\%                                  & 93.6\%                           \\
                         & RN50              & SatlasNet               & 92.9\%                                        & 93.8\%                            & 92.8\%                                  & 93.4\%                           \\
                         & Swin-B            & SatlasNet               & 93.6\%                                        & 85.3\%                            & 93.1\%                                  & 93.6\%                           \\
                         & Swin-T            & SatlasNet               & 91.3\%                                        & 94.0\%                            & 91.1\%                                  & 94.4\%                           \\
                         & Swin-B            & Supervised (ImageNet1k) & 92.0\%                                        & 92.8\%                            & 93.0\%                                  & 93.4\%                           \\
                         & Swin-T            & Supervised (ImageNet1k) & 92.6\%                                        & 92.5\%                            & 91.7\%                                  & 93.3\%                           \\
                         
    \midrule
\multirow{6}{*}{UNet}                        & MV2               & Supervised (ImageNet1k) & 94.4\%                                        & 94.5\%                            & 93.2\%                                  & 93.8\%                           \\
                        & RN50              & Supervised (ImageNet1k) & 93.7\%                                        & 94.7\%                            & 94.1\%                                  & 94.1\%                           \\
                        & RN50              & MoCov3 (ImagetNet1k)    & 92.0\%                                        & 94.2\%                            & 93.6\%                                  & 94.2\%                           \\
                        & RN50              & SeCo                    & 92.6\%                                        & 93.8\%                            & 92.8\%                                  & 94.4\%                           \\
                        & Swin-B            & Supervised (ImageNet1k) & 94.1\%                                        & 94.3\%                            & 93.2\%                                  & 94.1\%                           \\
                        & Swin-T            & Supervised (ImageNet1k) & 92.2\%                                        & 93.9\%                            & 93.6\%                                  & 93.3\%                           \\
                        \midrule
                        \midrule
    Average                        &                   &                         & 91.8\%                                        & 92.1\%                            & 91.3\%                                  & \textbf{92.7\%}                 \\
    \bottomrule
\end{tabular}
\caption{\textbf{Difference in performance across different loss functions used for training.} Binary cross entropy (bce) has the highest average performance across different settings. The other losses are adaptive max-pooling, Dice loss (dice), and centerline-aware loss (cldice).}
    \label{tab:loss-fns-explore}
\end{table*}

% \begin{figure*}[t]
%     \begin{center}
%     \includegraphics[width=0.8\linewidth]{figures/results_distribution_width_estimates.pdf}
%     \end{center}
%     % \vspace{-6pt}
%     \caption{\textbf{Distribution of river width estimates}. The y-axis represents the ground truth river widths derived from labeled PlanetScope images, while the x-axis represents the predictions across different models. Both Sentinel and SWOT models tend to overestimate the widths, while Landsat derived widths tend to underestimate the widths.
%     }
%     \label{fig:river-width-results-distrib-scatter}
%     % \vspace{-5mm}
% \end{figure*}

% \section{Reproducibility Checklist}

% \paragraph{Dataset availability in the data appendix.} We attach a partial dataset in the data appendix due to limitations on size. Upon publication, we will release a publicly available link to all the data introduced in this paper (and the corresponding code).

% \input{ReproducibilityChecklist}

\end{document}